\documentclass[10pt,twocolumn,letterpaper]{article}
\usepackage[numbers,sort&compress]{natbib}

\usepackage{iccv}
\usepackage{times}
\usepackage{epsfig}
\usepackage{graphicx}
\usepackage{amsmath}
\usepackage{amssymb}
\usepackage{booktabs}
\usepackage{makecell}
\usepackage{multirow}
\usepackage{enumitem}
\usepackage[accsupp]{axessibility}

\usepackage{xspace}
\usepackage[dvipsnames]{xcolor}
\usepackage{lipsum}

\newcommand{\backronym}{ChopNLearn\xspace}

\newcommand{\ob}{\texttt}
\newcommand{\st}{\texttt}

\usepackage[pagebackref=true,breaklinks=true,colorlinks,bookmarks=false,hyperfootnotes=false]{hyperref}

\usepackage[capitalize]{cleveref}
\crefname{section}{Sec.}{Secs.}
\Crefname{section}{Section}{Sections}
\Crefname{table}{Table}{Tables}
\crefname{table}{Tab.}{Tabs.}

\makeatletter
\newcommand{\dashrule}[1][black]{%
  \color{#1}\rule[\dimexpr.5ex-.2pt]{4pt}{.4pt}\xleaders\hbox{\rule{4pt}{0pt}\rule[\dimexpr.5ex-.2pt]{4pt}{.4pt}}\hfill\kern0pt%
}
\newcommand{\rulecolor}[1]{%
  \def\CT@arc@{\color{#1}}%
}
\makeatother
\iccvfinalcopy %

\def\iccvPaperID{1914} %

\ificcvfinal\fi

\begin{document}

\title{Chop \& Learn: Recognizing and Generating Object-State Compositions}

\author{Nirat Saini\thanks{First two authors contributed equally.}
\qquad
Hanyu Wang\footnotemark[1]
\qquad
Archana Swaminathan
\qquad
Vinoj Jayasundara
\qquad
Bo He\\[1ex]
Kamal Gupta
\qquad
Abhinav Shrivastava\\[2ex]
{University of Maryland, College Park}
}

\maketitle
\ificcvfinal\thispagestyle{empty}\fi

\begin{abstract}
   Recognizing and generating object-state compositions has been a challenging task, especially when generalizing to unseen compositions.  In this paper, we study the task of cutting objects in different styles and the resulting object state changes. We propose a new benchmark suite Chop \& Learn, to accommodate the needs of learning objects and different cut styles using multiple viewpoints. We also propose a new task of Compositional Image Generation, which can transfer learned cut styles to different objects, by generating novel object-state images. Moreover, we also use the videos for Compositional Action Recognition, and show valuable uses of this dataset for multiple video tasks. 
   Project website:  \textcolor{magenta}{https://chopnlearn.github.io}.   
\end{abstract}

\vspace{-0.2in}
\section{Introduction}
\vspace{-0.05in}
Objects often exist in different shapes, colors, and textures in the real-world. These visually discernible properties of objects, also known as states or attributes, can be inherent to an object (\eg, color) or be a result of an action (\eg, chopped). Generalization to unseen properties of objects remains an Achilles heel of current data-driven recognition models (\eg, deep networks) that assume robust training data available for exhaustive object properties. However, humans (and even animals) ~\cite{chomsky1993minimalist,arnold2006semantic}  can innately imagine and recognize a large number of objects with varying properties, by composing a few known objects and their states. This ability to synthesize and recognize new combinations from finite concepts, called \emph{compositional generalization} is often absent in modern deep learning models~\cite{lake2017building}.

Several recent works have been proposed to study composition in terms of the disentanglement of objects and the states in images~\cite{mit,ut,ge2,oadis} as well as videos~\cite{look_c,multi,nirat,joint,fathi,nachwa,gta}. A few works have attempted to improve open-world text-to-image generation models~\cite{ruiz2022dreambooth,gal2022image} for the task of compositional generation. However, current suite of datasets lacks either granular annotations for object states or enough data to study how object states evolve under different conditions. Therefore, measuring the compositional generalizability of these models on different tasks remains an open challenge.  

\begin{figure}[t]
\centering
\includegraphics[width=0.9\linewidth]{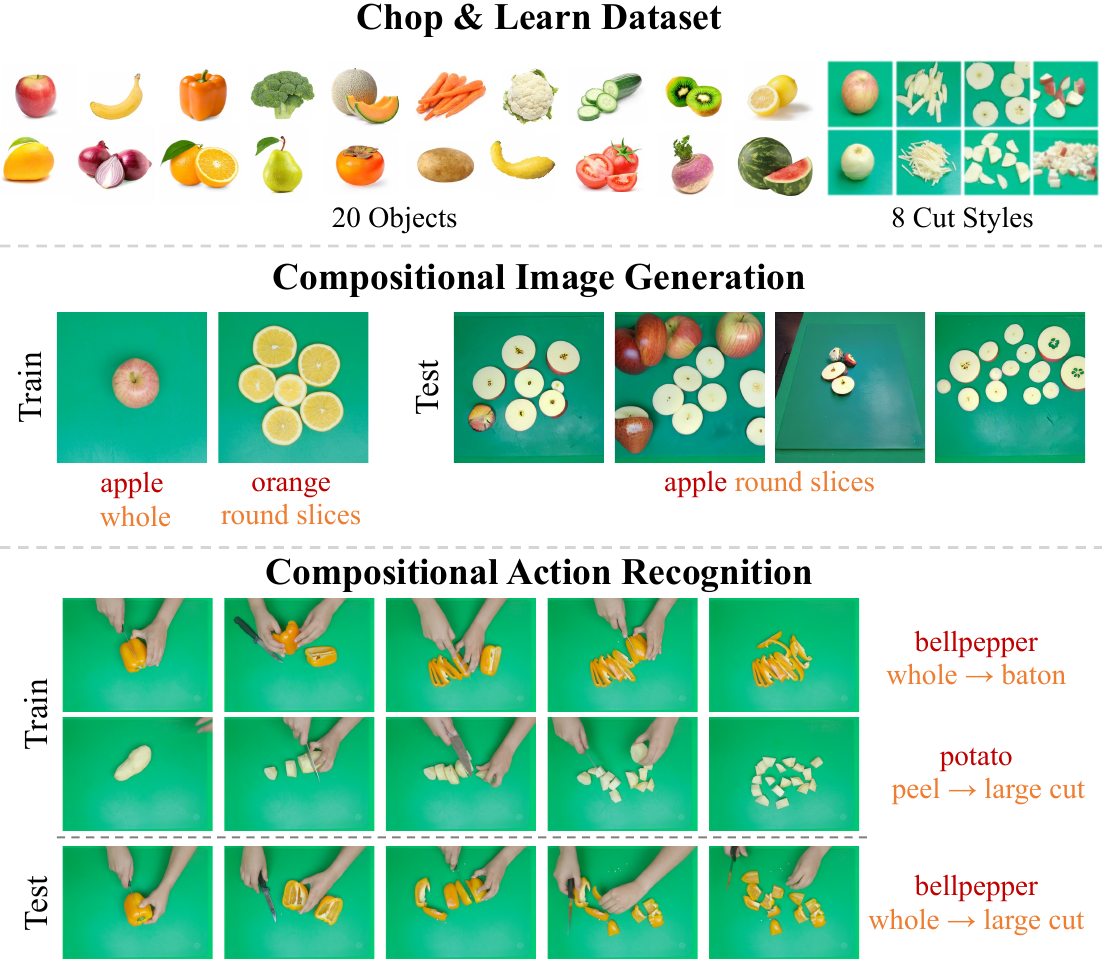}
\caption{We present \textbf{Chop \& Learn (\backronym}), a new dataset and benchmark suite for the tasks of Compositional Image Generation and Compositional Action Recognition. It consists of 1260 video clips and 112 object state combinations captured from multiple viewpoints for 20 objects and 8 cut styles. We also propose two new compositional tasks and benchmarks - (1) Image Generation: given training images of various {\color{Maroon}{objects}} in various  {\color{BurntOrange}{states}}, the goal is to generate images of unseen combinations of {\color{Maroon}{objects}} and {\color{BurntOrange}{states}}. (2) Action Recognition: training videos are used to recognize {\color{Maroon}{objects}} along with transition from {\color{BurntOrange}{state1}} $\rightarrow$ {\color{BurntOrange}{state2}}, to generalize on recognizing unseen object-state transitions.}
\label{fig:teaser}
\vspace{-0.18in}
\end{figure}

In this paper, we propose a new dataset, \textbf{Chop \& Learn} (\textbf{\backronym}) collected to support studying compositional generalization, the ability to recognize and generate unseen compositions of objects in different states. 
To focus on the compositional aspect, we limit our study to a common task in our daily lives -- cutting fruits and vegetables. When using different styles of cutting, these objects undergo different transformations and the resulting states are easily recognizable by humans. Our goal is to study how these different styles can be applied to a variety of objects for recognizing unseen object states. More specifically, we select \emph{twenty} objects and \emph{seven} commonly used styles of cuts (plus whole object) which results in object-state pairs with different granularity and sizes (\Cref{fig:teaser}). We collect videos of these objects being from \emph{four} different viewpoints, and label different object states in each video. Each style of cut changes the visual appearance of different objects in different ways. To study and understand object appearance changes, we propose two new benchmark tasks of Compositional Image Generation and Compositional Action Recognition, with a focus on unseen compositions.
 
The objective of the first task is to generate an image based on an (object, state) composition that was not seen during training. As shown in \Cref{fig:teaser}, during training, a generative model is provided with images of an (\ob{apple}, \st{whole}) as well as an (\ob{orange}, \st{round slices}). At the test time, the model has to synthesize a new unseen composition (\ob{apple}, \st{round slices}). We propose to adapt large-scale text-to-image generative models for this task. Specifically, by using text prompts to represent the object-state composition, we benchmark several existing methods such as Textual Inversion~\cite{gal2022image} and DreamBooth~\cite{ruiz2022dreambooth}. We also propose a new method by introducing new tokens for objects and states and simultaneously fine-tuning language and diffusion models. Lastly, we discuss the challenges and limitations of prior works as well as the proposed generative model with an extensive evaluation.
 
\begin{table}[t]
\caption{\textbf{Comparison with other video datasets.} This table highlights the distribution of the objects, states and compositions in different datasets. Obj. refers to objects, Comp. is compositions of objects and styles, $N$ refers to the number of compositions that have more than 10 samples, and Styles$^\ast$ refers to grouping of styles: instead of generic names like cut, chop, etc., we use 3 distinct styles (chop/dice, peel, grate) as styles. MIT-States$^{\dagger}$  is the only image-based dataset, the rest are video-based datasets. All these data numbers are for edible objects and cutting style actions from respective datasets. Our dataset has uniform distribution for each metric in the table, which makes it suitable for learning objects and their states.}
\label{tab:stats}
\centering
\footnotesize
\renewcommand{\tabcolsep}{2pt}
\renewcommand{\arraystretch}{1.2}
 \resizebox{\linewidth}{!}{
    \begin{tabular}{@{}lccccccccc@{}}
    \toprule
    \multirowcell{2}[-1.1pt][l]{Datasets} & \multicolumn{4}{c}{Total \# of} & \multicolumn{3}{c}{Avg.\ \# of Samples} 
    & \multirowcell{2}[-1.1pt][c]{$N$}
    & \multirowcell{2}[1.1pt]{\makecell{\# of\\ Views}}\\
    \cmidrule(r){2-5}
    \cmidrule(lr){6-8}
    & Samples &
Obj. &
\makecell{Comp.} &
\makecell{Styles$^\ast$} &
/Obj. &
\makecell{/Comp.} &
/Style &
& \\
    \midrule
MIT-States$^{\dagger}$ \cite{mit_states}  &	1676 &	27 &	52 &	4 &	62.07 & 	32.23 &	419 &	48 &	1 \\
Youcook2~\cite{youcook} &	714 &	160 &	313 &	3 &	7.3 & 	2.2 &	166.7 &	26 & 1 \\
VISOR~\cite{visor} &	301 &	58 &	122 &	3 &	5.2 & 	2.5 &	42.9 &	3 &		1 \\
COIN~\cite{coin} &	390 &	6 &	7 &	2 &	65 &	55 & 	195	& 6 &	 1 \\
Ego4D~\cite{ego4d} &	216 &	12 &	12 &	3 &	18.2 & 	18	& 54.5 &8 &	1 \\
50Salads \cite{Stein_2013_ACM} &	904 &	5 &	6 &	2 &	182 &	152 & 	457	& 6 & 1 \\
ChangeIt \cite{look_c} &	264 &	8 &	14 &	4 &	46.3 & 	26.4 &	96 & 14 	&	1 \\
CrossTask \cite{crosstask} &	1150 &	7 &	8 &	2 &	164.3& 	143.7 &	575 &	8 &	1 \\
Breakfast \cite{breakfast1} &	1055 &	3 &	4 &	2 &	351.7 & 	263.8 &	527.5 &	4 &	1 \\
\cmidrule{1-10}
\textbf{\backronym} &	\textbf{1260} &	\textbf{20} &	\textbf{112} &	\textbf{8} &	\textbf{74.2} & 	\textbf{11.8} &	\textbf{185.5} &	\textbf{112} &		\textbf{4} \\
    \bottomrule
    \end{tabular}}
\vspace{-0.15in}
\end{table}

In the second task, we extend an existing task of Compositional Action Recognition~\cite{something_else}. While the focus of prior work~\cite{something_else} is on long-term activity tracking in videos, we aim to recognize subtle changes in object states which is a crucial first step for activity recognition. By detecting the initial state and final object state compositions, our task allows the model to learn unseen object state changes robustly. We benchmark multiple recent baselines for video tasks on the \backronym dataset. 

Finally, we discuss various other applications and tasks that can use our dataset in image and video domains. To summarize, our contributions are threefold:
\begin{itemize}[leftmargin=*,itemsep=0em]
    \item We propose a new dataset \backronym, consisting of a large number of images and videos of diverse object-state compositions with multiple camera views.
    \item We introduce the task of Compositional Image Generation, which goes beyond the common conditional image generation benchmarks, and focuses on generating images for unseen object and state compositions.
    \item We introduce a new benchmark for the task of Compositional Action Recognition,
    which aims at understanding and learning changes in object states over time and across different viewpoints.
\end{itemize}

\begin{figure*}
\vspace{-0.15in}
\centering
\includegraphics[width=0.9\linewidth]{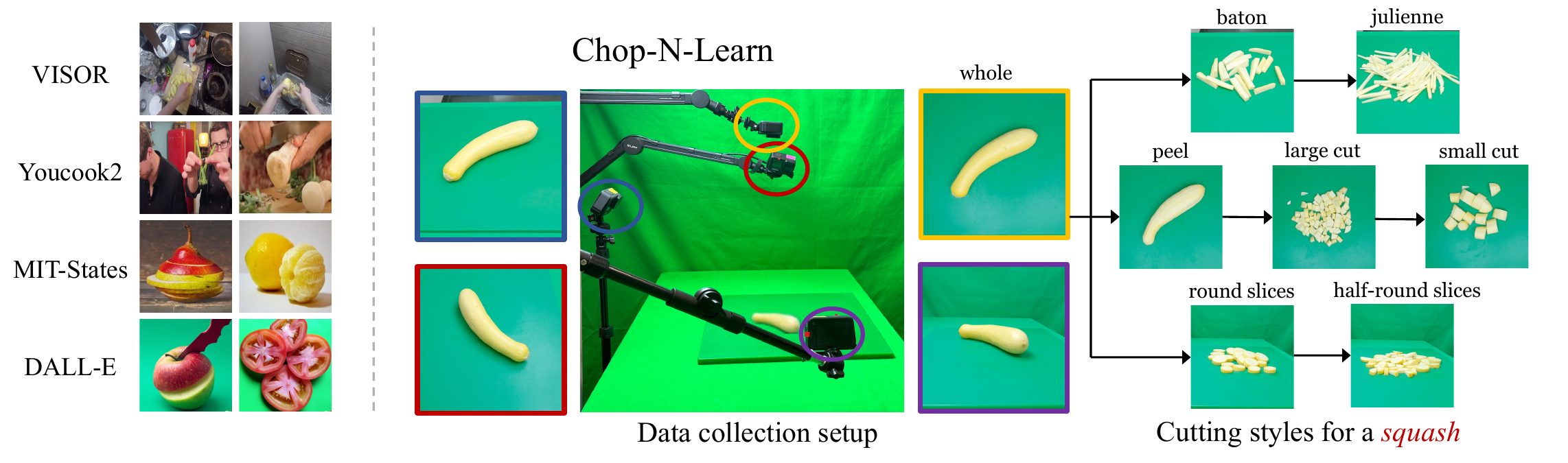}
\caption{\underline{Left}: We show examples of cutting styles from popular video datasets (VISOR~\cite{visor}: chop and peel potato, Youcook2~\cite{youcook}: chop broccoli, peel radish), image dataset (MIT-states~\cite{mit}:slice pear, peel orange) and generation pipelines (DALL-E~\cite{dalle}:baton cut apple, half round slices tomato). Most of these are either too noisy to capture subtle differences in objects or do not have the granularity of specific cutting styles. \underline{Center}: Our 4 camera setup captures videos of one object in 4 different views. \underline{Right}: We capture 8 styles of object states, which can be derived in a hierarchical manner from larger to small cuts. Each style is of different shape and granularity.}
\label{fig:data}
\vspace{-0.18in}
\end{figure*}%

\vspace{-0.15in}
\section{Related Work}
\vspace{-0.05in}

Object states or attributes have recently received significant attention for recognition tasks, in images and videos. Some of the common works and their dissimilarities with the proposed dataset are mentioned here.

\noindent \textbf{Attributes of Objects.} In the image domain, states are often referred to as attributes for Compositional Learning of attribute-object pairs. Attributes describe the visual properties of objects, such as shape, color, structure and texture. The common datasets used are MIT-states~\cite{mit}, UT-Zappos~\cite{ut}, COCO-attributes~\cite{ob5}, CGQA~\cite{ge} and VAW~\cite{vaw}. All of these datasets consist of web-scraped images of various types of objects (from furniture to shoes and clothes to food items), which makes the variety of states very diverse. Most of the prior works~\cite{ao,tmn,learn,symnet,rel,ge,ge2,oadis, pham2022improving, shrivastava2012constrained} focus on attribute-object recognition tasks using compositional learning but do not expand to image generation tasks due to the diversity in background and attributes.  Some works in compositional zero-shot learning of attributes show visual disentanglement of attributes from objects~\cite{oadis,gen_1}, however, they only hallucinate compositions of unseen attribute-object pairs in the feature space, rather than the image space. Moreover, even newer large vision-language models such as CLIP~\cite{clip}, DALL-E~\cite{dalle} fail to capture the subtle attributes of objects which are visually discernible~\cite{chensun,meshry}. Therefore, the image generation task for objects with different attributes is still unexplored, which is a major focus of our work.

\noindent \textbf{States for Action Recognition.} Detecting object states and corresponding actions from videos is explored in supervised~\cite{joint,nachwa,fathi,nirat} and self-supervised manners~\cite{look_c,multi,temp}. While some works focus on recognizing actions using states~\cite{nachwa,fathi,nirat,joint}, others discover states as the future frames in the videos in~\cite{temp,tap}. Some works~\cite{look_c,multi} also detect the exact frames of state 1, state 2 and the action that causes transition from state 1 $\rightarrow$ 2. Another recent work 
 (Ego4D~\cite{ego4d}) also proposes new tasks like point-of-return state-change prediction for object state transition detection. Hence, object states so far have been used as a signal for detecting and localizing actions. 
 We focus on extending this understanding of states to generalize across different objects with limited seen object-state transition videos. 
\\ 
\noindent \textbf{Compositional Action Recognition.} In contrast to randomly assigning samples for training and testing,~\cite{something_else} presented a new task of Compositional Action Recognition. The premise of this task is: actions are split based on objects they apply on. During training, only a set of objects are seen corresponding to set of objects, while during testing, unseen object appear for seen action labels. Following studies~\cite{stlt,strg,safcar,hand,look_l} used relationship between objects and states bounding boxes to model the compositional aspect, where the evaluation is performed on how well the composition of unseen object and state is recognized. We propose a similar task,
where videos are trained on seen compositions and tested on unseen compositions. \\
\noindent \textbf{Comparison with existing Datasets.} The existing image datasets such as MIT-states~\cite{mit}, UT-Zappos~\cite{ut}, COCO-attributes~\cite{ob5}, CGQA~\cite{ge} and VAW~\cite{vaw}, are not suitable for image generation tasks for two reasons:  1) there are very few transferable objects and attributes, 2) the images are web-scraped and very diverse with varied background. Due to this, generative models latch on background details rather than understanding subtle changes in objects. In video domain, there have been various video datasets with procedural and kitchen activities that capture object and state transformations, such as Epic-Kitchens~\cite{epic} with object and hand bounding box annotation version VISOR~\cite{visor}, Youcook2~\cite{youcook}, Ego4D~\cite{ego4d}, COIN~\cite{coin}, HowTo100M~\cite{howto}, Breakfast~\cite{breakfast1}, 50Salads~\cite{Stein_2013_ACM}, CrossTask~\cite{crosstask} and ChangeIt~\cite{look_c}. There are a few common problems across these datasets: (1) Most of these datasets lack annotations for the granularity of cutting styles. The styles labeled are \st{cut}, \st{chop}, \st{slice}, \st{dice}, \st{peel}, \st{grate}, \st{julienne}, which only comprises of three broader styles of transformations, \ie \st{chop/dice}, \st{peel} and \st{grate}. 
(2) The compositions of different objects and states are highly skewed and similar to image datasets. Some datasets have a long-tail distribution of objects, which can make it challenging for models to learn per-object-based states when there is only one sample available in the dataset. And lastly (3), the frames are noisy with lots of objects and attributes that object states changes are harder to capture (as shown in left side of \Cref{fig:data}). 
For most datasets, the ground truth is also not annotated for object detection, which makes it even harder to look for object of interest. Using an object detector to remove the background is an option, however with deformable objects, most Faster-RCNN~\cite{frcnn} based object detectors fail to capture the object itself, and latch onto smaller pieces instead. In Table~\ref{tab:stats}, we show statistics of data available in different datasets. The \# of clips from other datasets that has granular annotations of object-state pairs and can be used for compositional tasks. For instance, COIN~\cite{coin} has 180 categories with 10000 videos, but clips that have cutting styles as labels were only 390. Further, these clips only cover cut/peel actions, and cannot be categorized further based on granularity and shape of pieces. Our proposed dataset \backronym is designed to capture various objects and their cut styles, with uniformly distributed samples for 20 objects and 8 styles (including whole, 7 other cut styles~\Cref{fig:data}). 

\vspace{-0.05in}
\section{Chop \& Learn}\label{ref:dataset}
\vspace{-0.01in}

Our main objective with Chop \& Learn (\backronym) is to understand and learn granular object states, specifically styles of cuts which can be applied to diverse variety of objects. With this in focus, we collect object state transition videos, as well as images of object in various states, with 4 different camera views (\Cref{fig:data}). We discuss the design choices and motivation below.

\begin{figure}
\centering
    \includegraphics[width=\linewidth]{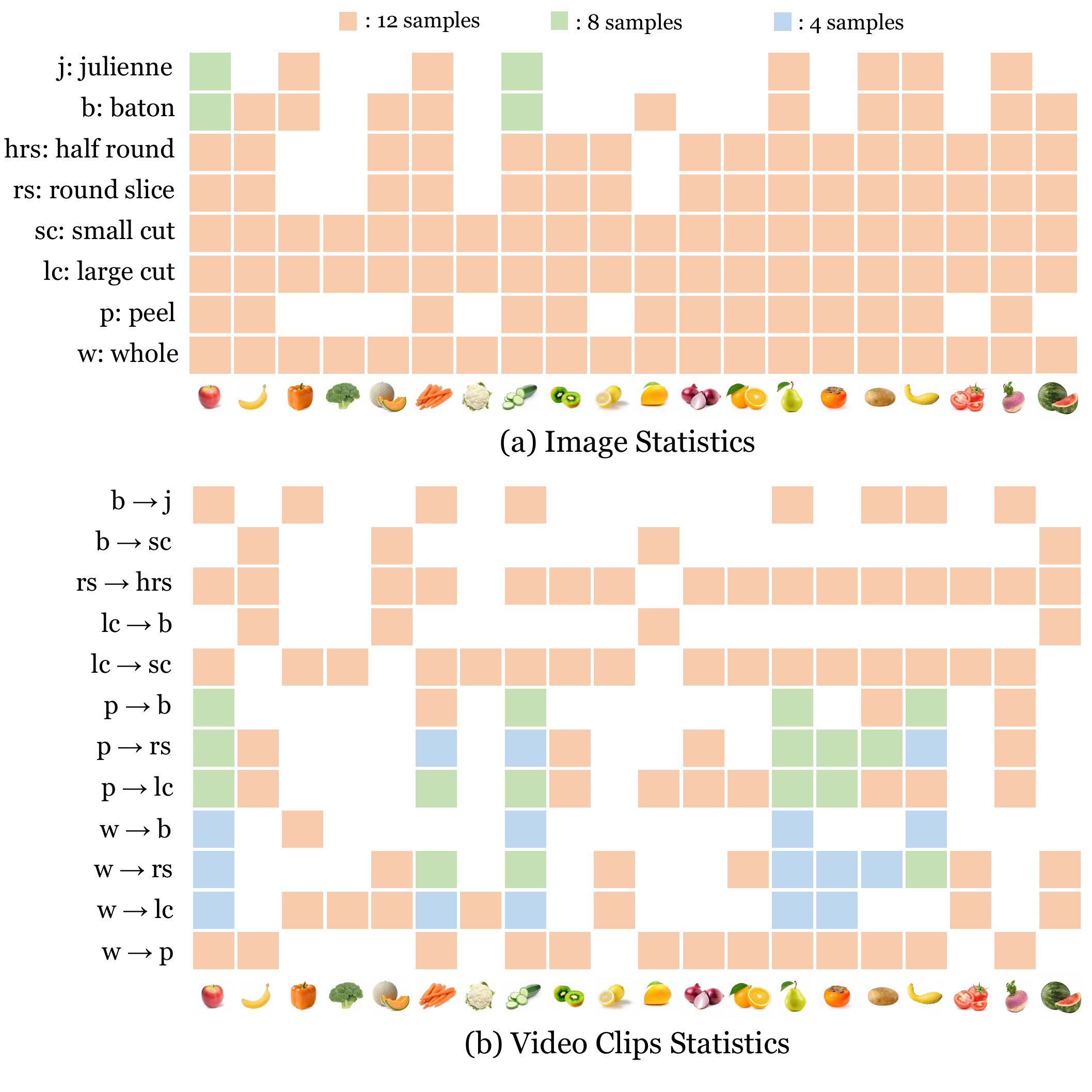}
    \vspace{-0.2in}
\caption{\textbf{Statistics for \backronym:} We show the number of samples for each object-style composition in a color-coded manner: orange represents 12 samples, green represents 8 samples and blue represents 4 samples.}
\label{fig:stats}
\vspace{-0.15in}
\end{figure}

\begin{figure*}
\centering
\includegraphics[width=0.85\linewidth]{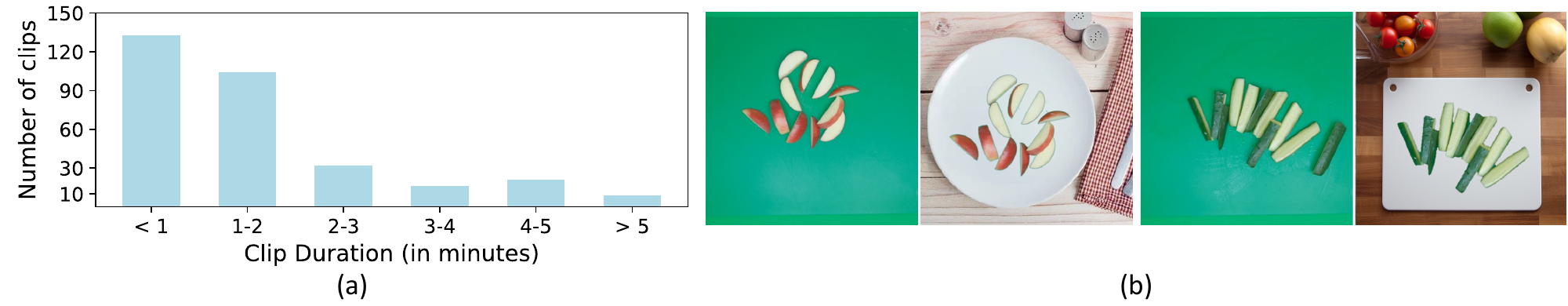}
\caption{\textbf{(a)} The clip length distribution for one camera (315 unique clips). \textbf{(b)} Preliminary results of using green screen to augment the dataset with different backgrounds. We continue to improve the transfer results by adding shadows and background matting.
}
\label{fig:gs}
\vspace{-1.3em}
\end{figure*}

\subsection{Design Choices}
\label{sec:data_collection}
\vspace{-0.05in}

\noindent \textbf{Selection of States (styles of cuts).} Fruits and vegetables are commonly cut in specific styles based on the need of the recipes. For instance, for eating an apple, we slice it in relatively large pieces while for using it in a pie, we might cut smaller or round slices of it.  We select 8 common styles of cuts, \ie, \st{large cut}, \st{small cut}, \st{baton}, \st{julienne}, \st{round slices}, \st{half round slices}, \st{peel}, and \st{whole} for our study. These are the most common styles of cuts for vegetables and fruits, which do not require any additional training to learn apart from common kitchen operation and knife handling skills. These styles of cuts can also have similarities with respect to shapes, yet are different in granularity. For example, \st{baton} (french-fries style cut) and \st{julienne} are similar in shape (long pieces), but \st{julienne} is more finely cut than \st{baton}. Similarly, \st{large cut} is a coarser version of \st{small cut}, and \st{half round slice} is one step from \st{round slices} (as shown in \Cref{fig:data}). We also have annotated the states \st{whole} and \st{peel}, which are the base states of objects.\\ 
\textbf{Selection of Objects.} We want to learn to transfer styles of cuts to different objects. To ensure consistency in transfer, we also consider the base state, \ie, \st{whole} state of objects.  For instance, it is hard to visualize \st{large cut} of carrots, if the seen data only includes rounder objects like \ob{oranges}. Hence, we consider some fruits and vegetables with similar colors, textures and shapes to include consistency across visual similarities after chopping. In this study, we used seasonal fruits and vegetables categorised on the basis on their shapes, colors and textures: round small objects: [\ob{apple}, \ob{pear}, \ob{mango}, \ob{potato}, \ob{turnip}, \ob{onion}, \ob{kiwi}], citrus fruits [\ob{lemon}, \ob{orange}], flower-like textured objects: [\ob{cauliflower}, \ob{broccoli}] , larger round objects: [\ob{cantaloupe}, \ob{watermelon}], textured from inside objects: [\ob{bellpepper}, \ob{tomato}, \ob{persimmon}], and long objects: [\ob{cucumber}, \ob{carrot}, \ob{squash}, \ob{banana}]. This consists of 10 fruits and 10 vegetable items, with at least one pair of similar objects presents in the dataset.

\noindent\textbf{Related Groups.} One of the key aspects of this dataset is transferability of cut styles to a variety of objects. We set up some constraints and create related groups for objects and styles. These related group enable us with structural and visual style transfer abilities. If an object is seen from related group $A$ with a particular style, we should be able to transfer that style to another object from the same related group $A$ and vice-versa. In other words, we group sets of objects and cut styles which are visually similar (based on color, shape and texture) together to create related groups for objects and states separately.
For states, we combine [\st{baton}, \st{julienne}], [\st{round slices}, \st{half-round slices}], and [\st{large cut}, \st{small cut}] together as related groups. 
For objects, we define seven groups with related objects: [\ob{apple}, \ob{pear}, \ob{mango}], [\ob{lemon}, \ob{orange}], [\ob{cauliflower}, \ob{broccoli}], [\ob{cantaloupe}, \ob{watermelon}, \ob{kiwi}], [\ob{bellpepper}, \ob{tomato}, \ob{persimmon}], [\ob{potato}, \ob{turnip}, \ob{onion}], and [\ob{cucumber}, \ob{carrot}, \ob{squash}, \ob{banana}].\looseness=-1

\vspace{-0.05in}
\subsection{Data Collection Setup} 
\vspace{-0.05in}
We collect data using four GoPro cameras \cite{gopro} positioned at different angles, with three participants (\Cref{fig:data}). We use a green screen and green chopping board for minimum distraction in the background, such that the objects and their cut pieces are easily segmented for each view.

\noindent \textbf{Granularity of styles.} For ease and consistency across participants, the size of cut pieces can be defined as the shape and ratio of one piece with respect to the whole object. For more details, please refer to the appendix. Given a set of $n$ states and $m$ objects, we can have at most $m \times n$ compositions. However, our dataset does not include some compositions which are not commonly found in real world. For instance, due to the texture of \ob{onions}, it is not feasible to cut \ob{onions} in \st{baton} or \st{julienne} style, since the layers of the \ob{onion} do not stay intact, so we do not have a sample of [\st{baton}, \ob{onion}]. 

\noindent \textbf{Video Recording.}
We primarily collect video data, and derive state change frames from long videos. Each video consists of 2-3 object states, which are annotated while data collection process using the highlight features of GoPros. For synchronizing across different cameras, we initially start with a clapper to make a clap sound for indicating the beginning of the video. Then, we highlight the frames in one of the GoPro as the first/initial state. The participant then walks up the object and starts cutting the object. After the object is cut in one style, the participant steps back and we highlight another frame as the next state. The participant performs at least 2 styles of cut in each video, which can be done consecutively. For instance, we can first cut an object with \st{large cuts}, and then do \st{small cuts} subsequently. The video ends with another clap for the end of video detection and synchronization across different cameras. Henceforth, we collect video data along with annotated states for each participant, without extra effort of annotations. More details and statistics of dataset are shown in \Cref{fig:stats}. Average video clip length (one state change for an object) is 1m40s. The distribution is shown in Fig.~\ref{fig:gs}(a).

\begin{figure*}[!t]
\vspace{-0.15in}
\centering
    \includegraphics[width=0.95\linewidth]{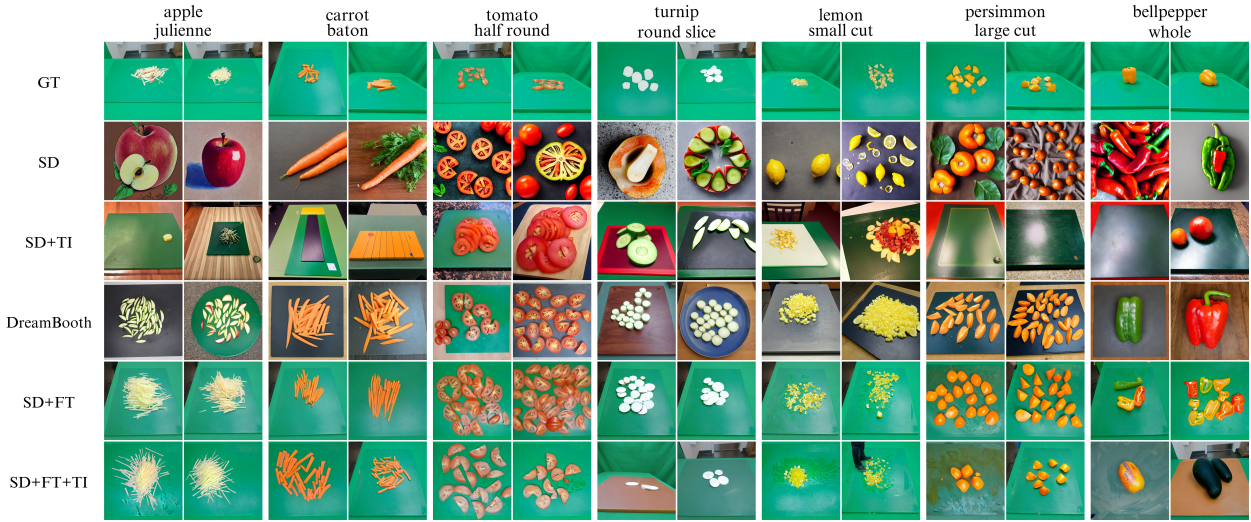}
    \caption{\textbf{Compositional Generation Samples.} Ground Truth (GT) real images are shown in the first row for reference. Seven object-state combinations in the test set are displayed, each with two generated samples for each method. Please zoom in to see details.} 
\label{fig:comp_test}
\vspace{-0.18in}
\end{figure*}
\vspace{-0.1in}
\section{Compositional Image Generation}
\begin{table}[t]
    \centering
        \caption{\textbf{Compositional generation evaluation.} FID, user scores, and classifier scores of various generative models. User Realism is on a scale of 1-5. ($\star$) denotes that accuracies are evaluated on a seen data split. \textbf{Bold} represents the best result.}
        \resizebox{\linewidth}{!}{
        \begin{tabular}{@{}l|c|c|cc|cc} 
            \toprule
            Method & Patch & User  &  \multicolumn{2}{c|}{Classifier Acc. (\%)} &  \multicolumn{2}{c}{User Acc. (\%)}\\
            & FID $\downarrow$ & Realism $\uparrow$ & Object $\uparrow$  & State $\uparrow$  & Object $\uparrow$  & State $\uparrow$ \\
            \midrule
            Real Images & - & 4.65 & 87.5$^\star$ & 92.0$^\star$ & 73.6 & 84.0 \\
            \midrule
            SD          & 178.0 & 3.41 & \textbf{73.1} & 27.9 & \textbf{81.6} & 28.8 \\
            SD+TI       & 145.0 & 2.58 & 23.6 & 37.7 & 21.6 & 43.2 \\
            DreamBooth  & 139.9 & 3.56 & 53.5 & 74.2 & 61.6 & 72.8 \\
            SD+FT       & 88.9  & \textbf{3.78} & 70.5 & 67.7 & 72.0 & 65.6 \\
            SD+FT+TI    & \textbf{82.2}  & 3.47 & 67.8 & \textbf{81.4} & 67.2 & \textbf{79.2} \\
            \bottomrule
        \end{tabular}
        }
        \label{tab:eval_metrics}
        \vspace{-1em}
\end{table}
Large-scale deep generative models~\cite{rombach2022high,ramesh2022hierarchical,saharia2022photorealistic} trained on open-world big datasets have made significant breakthroughs in image generation in the last couple of years. These models, are typically conditioned using a text encoder and also support tasks such as zero-shot image generation, inpainting, image editing, and super-resolution without explicit training on these tasks. However, the performance of these models significantly degrades when it comes to compositional generation~\cite{dalle2}. Our dataset, consisting of 112 real-world object and state combinations, is well-suited to test the compositional capabilities of generative models.

\smallskip\noindent
\textbf{Task Description.} The goal of the task is to either train from scratch or fine-tune an existing generative model using the (object, state) pairs provided in the training, and generate images from unseen compositions. We consider all 20 objects, each object captured in up to 7 different states, \ie, all the states excluding \st{peel}. We split the (object, state) combinations into a training set consisting of 87 combinations and a test set consisting of 25 combinations. The training set covers all objects and states used in our dataset, but it does not overlap with the test set in terms of (object, state) combinations. In other words, for each combination of object and state present in the test set, the training set includes exactly one of either the object, or the state, but not both. We also ensure that for each (object, state) combination $(o, s_i)$ in the test set, there exists a combination $(o, s_j)$ in the training set, where $s_i$ and $s_j$ belong to the same state related group defined in Section~\ref{sec:data_collection}.
This setting ensures that all object and state information are available in the training set. 
Each combination in our dataset has 8-12 images, resulting in a total of 1032 images in the training set and 296 images in the test set.  The exact split is provided in the appendix along with some examples.
\subsection{Methods} 
\noindent
\textbf{Stable Diffusion. (SD)}
We evaluate a popular open-source text-to-image generative model Stable Diffusion (SD)~\cite{rombach2022high}. For details on the SD, refer to the original work~\cite{rombach2022high}. Here we briefly describe the sampling process. Diffusion models generate an image from Gaussian noise via an iterative denoising process. SD uses classifier-free guidance~\cite{ho2022classifier} for sampling. This means given a text prompt $\mathbf{c}$, we encode the prompt using CLIP's text classifier~\cite{clip} and recursively update a Gaussian noise sample with 
\begin{align}
    \omega \boldsymbol{\epsilon}_\theta(\mathbf{x}_t, \mathbf{c}) + (1-\omega)\boldsymbol{\epsilon}_\theta(\mathbf{x}_t)
\label{eq:guidance}
\end{align}
\noindent
where $\mathbf{x}_t$ is the denoised sample at the time step $t$ and $\boldsymbol{\epsilon}_\theta$ is SD. With each time step, we try to move the denoised sample using the guidance provided by the text prompt. The strength of the guidance is defined by $\omega$. 

As our first baseline approach, we sample zero-shot images from SD with a text prompt ``An image of $o_i$ cut in $s_j$ style", where $o_i$ is the $i^{th}$ object and $s_j$ is the $j^{th}$ state of the object. Zero-shot generation with a pre-trained SD model doesn't work as intended as shown in \Cref{fig:comp_test}, and the generated images often perform poorly in capturing the object state.
Several recent works have shown that it is possible to extend models such as SD to achieve high-quality customized generations~\cite{gal2022image,ruiz2022dreambooth,zhang2023adding}. We evaluate several methods that have been proposed for compositional generation in the recent literature. We also propose a simple yet strong baseline by fine-tuning a Stable Diffusion (SD) model~\cite{rombach2022high} along with textual inversion. 

\noindent
\textbf{SD + Textual Inversion (TI).} Textual Inversion~\cite{gal2022image} introduces new tokens in the vocabulary and optimizes their embedding from the given images keeping SD frozen. We adapt the method for our task by introducing new tokens for the objects $\{o_i\}$ and the states $\{s_j\}$, and jointly optimize the embeddings of ${\{o_i\}  \cup \{s_j\}}$ by providing (image, prompt) pairs from our training data. 
As before, the prompt is simply constructed as  ``An image of $o_i$ cut in $s_j$ style". %

\noindent
\textbf{DreamBooth.} Next, we adapt DreamBooth~\cite{ruiz2022dreambooth}, which fine-tunes the diffusion model along with the state-specific tokens. In our experiments, we fine-tune one model for each state in the dataset, where only the state token is learned. Original DreamBooth optimizes the diffusion loss as well as a prior preservation loss~\cite{ruiz2022dreambooth}. We observed that the latter significantly deteriorates the performance thus we skip it.

\noindent
\textbf{SD + Fine-tuning (FT).} We also fine-tune SD. In this baseline, only the parameters in the UNet of the diffusion model are optimized while keeping the text encoder fixed. %
\noindent
\textbf{SD + TI + FT.} Finally, we combine SD fine-tuning and Textual Inversion~\cite{gal2022image}. Specifically, on top of our SD + Fine-tuning baseline, we also adapt Textual Inversion by introducing new object tokens and state tokens and optimizing their embeddings along with the UNet parameters. 

\vspace{-0.05in}
\subsection{Evaluation} 
\label{sec:gen_eval}
\vspace{-0.05in}
\noindent We use both qualitative and quantitative measures to evaluate the capabilities of different methods. This section explains the details of different evaluation metrics we used:

\noindent
\textbf{Patch FID.} Fr\'echet Inception Distance (FID)~\cite{heusel2017gans} is a commonly used metric to assess the quality of generative models. Given a set of real images and a set of generated images, FID compares the mean and std of Inception-v3 features of the two sets.
For each composition and generative model, we compute patch FID using all real and 16000 generated patches, and report the average number for the test pairs. We hypothesize that using patch FID gives more weight to the object-state patches, rather than the whole image, which includes almost 50\% background pixels. We further calculate the lower bound for patch FID score by computing it between two sets of real images. Any score lower than that for this dataset can be disregarded as irrelevant. The determined lower bound for the patch FID score is 37.2. 

\noindent
\textbf{Object/State Accuracy using a Classifier.}
To evaluate the correctness of objects and states in the generated images, we train a classifier on real images for classifying objects and states independently. This classifier is built on top of CLIP-ViT-B/32~\cite{clip}. Classification logits are obtained by computing the cosine similarity between the image embedding and text embeddings of all possible state labels or object labels. To ensure the reliability of the classifier's results, we train it on the training set from a different dataset split, where all (object, state) combinations are present.

\noindent
\textbf{User Study.}
We conducted a user study to evaluate the generated images. We took images from the test set as well as samples from our generative models and present them to 30 users. Each user was presented with 25 distinct images, randomly sampled with an even distribution from our models and the test set. After giving a tutorial to the users about the different objects and states present in our experiments, the users were asked to choose an appropriate object name and state label, as well as rate the image for realism on a scale of 1-5. We report the object and state accuracies as well as realism score in \Cref{tab:eval_metrics}. The details of our user study design can be found in the appendix.

\renewcommand{\thefootnote}{\fnsymbol{footnote}}

\begin{table*}[t]
\centering
\caption{\textbf{Compositional action recognition results.} ``Start/End" denote the prediction results for the initial and the final state composition with the corrected object type. \textbf{Bold} and \underline{underline} represent the top-1 and top-2 results.} 
\label{tab:res_vid}
\footnotesize
\renewcommand{\tabcolsep}{3pt}
\resizebox{0.9\textwidth}{!}{
    \begin{tabular}{@{}llcccccccccccccc@{}}
    \toprule
    & & \multicolumn{4}{c}{Split 1}
    & & \multicolumn{4}{c}{Split 2}
    & & \multicolumn{4}{c}{Split 3}\\
    \cmidrule{3-6}
    \cmidrule{8-11}
    \cmidrule{13-16}
    & & \multicolumn{2}{c}{Start} & \multicolumn{2}{c}{End} 
    & & \multicolumn{2}{c}{Start} & \multicolumn{2}{c}{End}
    & & \multicolumn{2}{c}{Start} & \multicolumn{2}{c}{End}\\
    \cmidrule{3-4}
    \cmidrule{5-6}
    \cmidrule{8-9}
    \cmidrule{10-11}
    \cmidrule{13-14}
    \cmidrule{15-16}
    Model &  Features & acc@1  & acc@3 & acc@1  & acc@3 & & acc@1  & acc@3 & acc@1  & acc@3 && acc@1  & acc@3 & acc@1  & acc@3  \\
    \midrule
    AvgPool & I3D~\cite{i3d} & 9.5 & 23.7 & 4.7 & 14.2 & & 8.3 & 21.9 & 5.2 & 19.8 & & 15.9 & 28.5 & 4.8 & 22.3\\
    LSTM~\cite{hochreiter1997long} & I3D~\cite{i3d} & 14.2 & 36.2 & 5.7 & 29.8 & & 12.5 & 29.2 & 6.2 & 26.0 & & 17.5 & 34.9 & 6.3 & 23.7\\
    Transformer~\cite{vaswani2017attention} & I3D~\cite{i3d} & 23.7 & 49.0 & 10.9 & 44.3 && 27.5 & 46.2 & 14.6 & 44.2 && 20.6 & 42.9 & 11.1 & 44.4\\
    \midrule
    AvgPool & MIL-NCE~\cite{mil} & 11.1 & 31.6 & 4.8 & 28.4 & & 9.4 & 17.7 & 5.2 & 13.5 & & 14.2 & 41.4 & 12.8 & 41.4\\
    LSTM~\cite{hochreiter1997long} & MIL-NCE~\cite{mil} & 15.9 & 36.5 & 6.4 & 36.6 & & 11.9 & 36.7 & 9.8 & 36.7 & & 18.9 & 39.6 & 8.0 & 25.4\\
    Transformer~\cite{vaswani2017attention} & MIL-NCE~\cite{mil} & \underline{50.9} & 85.7 & 47.7 & 76.2 && \bf 56.2 & 82.3 & \underline{52.7} & 88.5 && 41.1 & 74.6 & 42.9 & 77.7\\
    \midrule
    STLT~\cite{stlt} & -- & 2.8 & 15.5 & 1.4 & 8.4 & & 1.4 & 13 & 1.4 & 11.6 & & 4.2 & 14.1 & 1.4 & 11.3\\
    Transformer~\cite{vaswani2017attention} & R3D~\cite{r3d} & 45.1 & \underline{85.9} & \underline{52.1} & \underline{85.9} & & \underline{55.1} & \underline{94.2} & \bf 58.0 & \underline{92.8} & & \underline{59.1} & \underline{85.9} & \underline{56.3} & \underline{85.9}\\
    CAF~\cite{stlt} & R3D~\cite{r3d} & \bf 53.5 & \bf 88.7 & \bf 57.8 & \bf 88.7 & & \underline{55.1} & \bf 95.7 & \bf 58.0 & \bf 95.7 & & \bf 62.0 & \bf 93.0 & \bf 63.4 & \bf 93.0\\
    \bottomrule
    \end{tabular}
}
\vspace{-0.15in}
\end{table*}
\subsection{Results and Discussion}
\label{sec:results}
\vspace{-0.05in}
\noindent
\textbf{Qualitative Results.}
\cref{fig:comp_test} displays the generated images from various methods for seven (object, state) combinations in the test set. The first row of the figure exhibits the ground truth real images for reference. We observe that vanilla SD often generates correct objects in random states, while SD+TI frequently synthesizes images without displaying the object. DreamBooth performs better than SD+TI, but worse than a simple finetuning of SD. SD+FT and SD+FT+TI perform well in terms of state generation.

\noindent
\textbf{Quantitative Results.}
Table~\ref{tab:eval_metrics} displays the performance of all baseline methods evaluated according to the metrics outlined in \Cref{sec:gen_eval}. Assessing image realism is a crucial evaluation metric for generative models; however, defining and measuring it can be challenging. Note that the patch FID values and user realism ratings do not align well. This is due to the disparity between the distribution of images in our dataset and that of typical occurrence of those objects in the real world.
The patch FID metric measures the similarity between the generated images with those in our dataset, instead of the ones most typical in real world. In particular, our results indicate that SD achieves the worst patch FID score since it has not encountered our dataset before, whereas its user realism rating is more satisfactory. 
SD+TI has the lowest user realism rating and a poor patch FID score, which suggests that only training object/state embeddings is inadequate for generating high-quality images. 
DreamBooth receives a good user realism rating but a poor patch FID, indicating that the images it generates are realistic but not very similar to those in our proposed dataset. Finally, fine-tuning via both SD+FT and SD+FT+TI achieve better results for patch FID and user realism.

We next evaluate the accuracy of objects and states in generated images. It is worth noting that the classification task on our dataset is intrinsically difficult, which leads to imperfect user accuracy on real images.
In general, the accuracy scores from classifier closely align with one from users, indicating that the proposed classifier is suited for evaluating compositional generation.

Our results show that SD achieves the best object accuracy but the worst state accuracy. This is possibly due to the lack of state variations in most existing large image datasets.
SD+TI is the worst performer due to its limited learning capacity. On the other hand, DreamBooth, SD+FT, and SD+FT+TI attain better state accuracy. Among them, DreamBooth's object accuracy is slightly worse as it is particularly trained for states. SD+FT achieves high object accuracy, and SD+FT+TI attains the best state accuracy with the help of fine-tuning and textual inversion together.

\noindent
\textbf{Green Screen Removal.} 
One of the main challenges for understanding fine-grained object-state pairs with existing datasets such as MIT-states~\cite{mit_states} is diverse backgrounds. 
Using them for training often leads to the model latching on to unwanted background details and missing out on the state understanding. Hence, we collected \backronym with a clean green screen background for the benchmark tasks. While we acknowledge the limitations it poses to our trained models, we highlight that the green screen can potentially enhance our ability to generalize to diverse scenes. This can be achieved by segmenting out images and placing various backgrounds, along with scaled and rotated object-state images (Figure~\ref{fig:gs}). As a proof-of-concept, we train a SD+FT+TI model on background-augmented images, and report the Patch FID, classifier object accuracy and state accuracy in \cref{tab:no_gs}. Note that here we employ a newly trained classifier that uses background-augmented images, and the patch FID scores are also computed based on these images. We further reference the lower bound of the patch FID as defined in Section~\ref{sec:gen_eval}. Due to the complex backgrounds introduced, the object accuracy and the patch FID of the new model are slightly compromised. However, it maintains a high and even improved state accuracy. This demonstrates the potential of the background-augmented \backronym in enhancing fine-grained compositional image generation.

\begin{table}[htbp]
  \centering
  \vspace{-0.1in}
  \caption{ \textbf{Green screen removal evaluation.} Both rows employ the SD+FT+TI but are trained using images with varying backgrounds. Classifiers specific to each dataset are trained to assess Classifier Acc. Validation images used to calculate Patch FID differ between the two rows. Patch FID Lower Bound is computed by evaluating the patch FID on one-half of the validation images relative to the other half. For further details, refer to Section~\ref{sec:results}. }
  \label{tab:baseline_results}
  \resizebox{0.4\textwidth}{!}{
    \begin{tabular}{@{}l|cc|c|c@{}}
      \toprule
      Data & \multicolumn{2}{c|}{Classifier Acc. (\%)}   & Patch FID $\downarrow$ & Patch FID \\
      Background & Object $\uparrow$ & State $\uparrow$ & & Lower Bound \\
      \midrule
      Green Screen & 67.8 & 81.4 & 82.2 & 37.2 \\
      Various & 46.3 & 82.3 & 133.6 & 46.4 \\
      \bottomrule
    \end{tabular}
  }
  \label{tab:no_gs}
  \vspace{-0.22in}
\end{table}

\vspace{-0.07in}
\section{Compositional Action Recognition}
\vspace{-0.05in}
Human actions often change object states and different objects can have diverse visual transitions even when subjected to the same action type.
To investigate this problem in a more intuitive manner,~\cite{something_else} introduced a new task of compositional action recognition, which targets at improving the robustness of models to handle the same actions with different objects involved.
For example, given an action of `taking something out from something', the model is trained on a limited set of objects and is tested on unseen types of objects to access its generalizability. 
Hence, despite the same underlying action, the object and visual features can be quite diverse. 
Similarly, the composition of the same action with different object types can look very distinctive. 
For instance, although cutting an \ob{carrot} and a \ob{apple} require similar knife movements, the resulting visual changes are distinct, with the former changing from a \st{whole} \ob{apple} to a \st{peeled} \ob{apple}, and the latter changing from a \st{whole} \ob{carrot} to a \st{peeled} \st{carrot}.
Therefore, we propose to use our dataset for the task of compositional action recognition, which can also be referred to as Compositional Zero-Shot Action Recognition, as the compositions of objects and states are unseen during training.  \\
\noindent \textbf{Task Description.} 
For this task, we consider each clip of a video as containing a single object with a single state transition. 
From the raw videos, which typically contain 2-3 transitions of object states per video, we segment the clips into isolated ones with only one transition. 
Examples of transitions include changing from a \st{whole} object to a \st{peeled} object or from a \st{peeled} object to a \st{baton} cut object. 
Similar to \cite{something_else}, we divide all object-final state compositions into two sets: seen compositions, which are used for training, and unseen compositions, which are used for testing. 
Following the approach used in the Compositional Image Generation task, we ensure that each object and state are seen at least once individually in the training set, but not together as a composition. 
The objective of the task is to predict the correct labels for the initial object-state composition $(o_i, s_j)$ and the final composition $(o_i, s_k)$, given a clip containing an object $o_i$ transitioning from an initial state $s_j$ to a final state $s_k$. 
Note that the clip is considered correctly classified only if both the object and state labels are correct for both the initial and final compositions.

\subsection{Dataset Splits}

We create 3 different dataset splits as follows (more details are in the Appendix). All splits have disjoint train, test and validation samples, and are created with different constraint combinations:
\begin{itemize}[leftmargin=*,itemsep=0em]
    \item \textbf{Split 1:} This split is a random selection of object-final state compositions with cross-view condition. We do not use any information from related groups. 
    \item \textbf{Split 2:} In this split, we use related group information for states, along with cross-view. based on related groups, if \st{baton} \ob{carrots} is seen in training set, then \st{julienne} \ob{carrots} can be part of test set. Since \st{baton} and \st{julienne} are part of the same related group, we can learn an object in one style and can generalize to another style from the same group in ~\Cref{sec:data_collection}.  
    \item \textbf{Split 3:} This split includes information from both related groups for states and objects. We want to ensure that even if an object is not seen in its related group, a similar object is seen in the related group. For example, if \ob{broccoli} is seen with \st{large cuts}, then \ob{cauliflower} with large or small cuts can be in the test set. 
\end{itemize}
Hence different splits represent different complexity levels for compositional action recognition.

\noindent \textbf{Evaluation.} We evaluate the accuracy of predicting both the initial and final compositions of objects and states in the test set.
Only when the object and state are both correct, it is counted as a correct prediction.
Specifically, we use two separate prediction heads for objects and states. 
We emphasize the need to evaluate composition as a whole, rather than just predicting the state, as the way an \ob{apple} is cut can differ significantly from the way a \ob{bellpepper} is cut.
Therefore, accurately recognizing both the object and state is crucial for tasks related to understanding and generating videos of object states. 
We also recognize the importance of top@3 accuracy, since object states can sometimes be visually similar, leading to confusion in detecting the correct composition. For example, \st{julienne} \ob{apple} can be visually very similar to \st{julienne} \ob{potato}.

\subsection{Results}
To evaluate our proposed method, we establish baselines using both traditional architectures and features for video action classification, as well as comparing with recent works in compositional action recognition. 
As shown in Table~\ref{tab:res_vid}, in the first section, we use pre-extracted I3D\cite{i3d} features and conduct experiments by comparing simple average pooling, LSTM, and multi-layer Transformer~\cite{vaswani2017attention} model. 
It shows that the Transformer model performs the best among these variants due to the great capacity of temporal modeling ability.
In the second section, we also experiment with more recent pre-trained features MIL-NCE~\cite{mil} along with transformer models, which outperforms I3D features. MIL-NCE~\cite{mil} features are pre-trained on HowTo100M~\cite{howto} with multimodal (RGB+narrations) setup, which is more robust for video downstream tasks.

\noindent In the final section of Table~\ref{tab:res_vid}, we employ the state-of-the-art compositional video recognition model proposed in~\cite{stlt} and use pseudo labels of bounding boxes for each hand and object, as there are no ground-truth hand and object trajectories available.
Specifically, the Spatial-Temporal Layout Transformer (STLT) \cite{stlt} takes in the spatio-temporal locations and class labels for each bounding box as input, uses positional embeddings to project coordinates and class labels into features, and adds transformer layers to model spatial-temporal relationships. 
However, without any appearance information, STLT achieves low performance on all metrics.
On the other hand, with the appearance features, which are extracted by inflated 3D ResNet50~\cite{kataoka2020would} (R3D), it can achieve much higher performances than STLT.
Finally, Cross-Attention Fusion (CAF) applies cross-attention~\cite{tan2019lxmert} to fuse the layout (STLT) and appearance (R3D) branch embeddings, achieving the best results.
It demonstrates that combining the layout and appearance information together can help predict object and state types more accurately.
\begin{figure}
\vspace{-0.15in}
\centering
    \includegraphics[width=\linewidth]{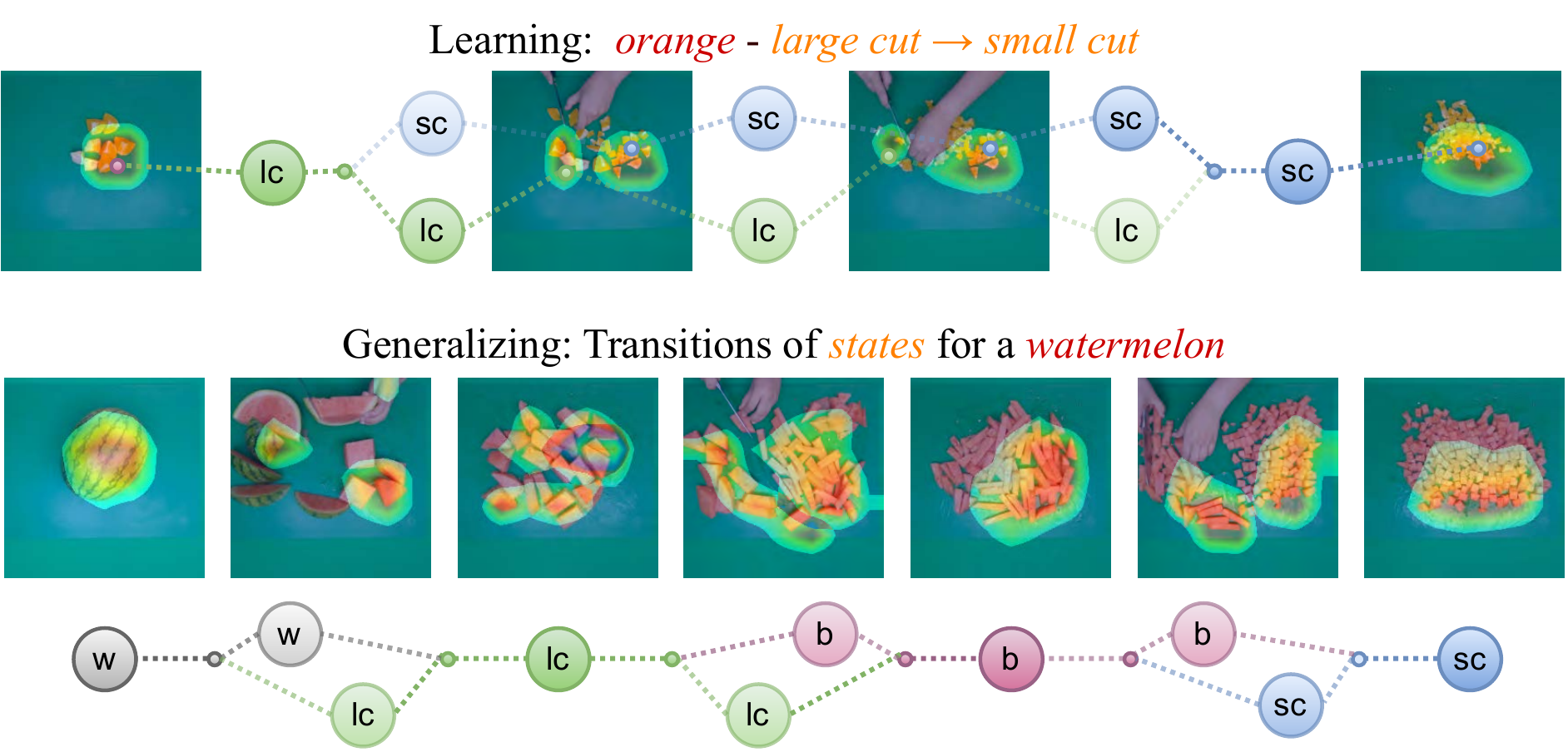}
    \caption{\textbf{Video parsing graph:} For a given video, we use Grad-CAM\cite{Selvaraju_2017_ICCV} on the intermediate frames to identify and visualize the class activation maps corresponding to the most salient states. Top: A training video clip has one transition of \ob{orange} from \st{large cut} $\rightarrow$ \st{small cut}. Bottom: We can learn single transitions from training data, to generalize transitions in a long video with multiple state changes and parse the video as a graph.}
\label{fig:test_parse_graph}
\end{figure}
\section{Discussion}
\vspace{-0.05in}
\label{sec:disc}

We discuss the potential future use of \backronym, while addressing the limitations and scope as well.

\noindent \textbf{Long-term Video Parsing.}
We use compositional state recognition to further understand the temporal dynamics~\cite{temp,bo_dnerv,bo_asm,a2summ} with the aid of a video parsing graph construction as previously explored in Ego-Topo~\cite{egotop} and VideoGraph~\cite{videograph}. Each clip in the training set has one state transformation (top example in \Cref{fig:test_parse_graph}). We visualize the class activation maps corresponding to the most salient intermediate state transitions with Grad-CAM~\cite{Selvaraju_2017_ICCV}, to learn the transition in each frame of the video for training data. This is illustrated as a graph for a training video. Having learned multiple single transformations, we can now extend this knowledge to understand long activities, with multiple transitions. As shown in \cref{fig:test_parse_graph}, we can learn state changes for \ob{orange} from \st{large cut} $\rightarrow$ \st{small cut} using our training clip. Given a long unseen video with multiple clips, we can construct a state-transition graph to represent changes in state for a \ob{watermelon}. %
Hence, by using an extensive array of videos, the process of learning transitions between individual states can be extended to encompass transitions between multiple states. This enables the creation of a self-supervised transition knowledge graph for comprehensive long-term video comprehension, as demonstrated in~\cite{temp,Wei_2018_CVPR}.

\noindent \textbf{Limitations.}
With advent of foundation models, few-shot generalization is an increasingly important task. In this work, we explore the potential of \backronym for the research in compositional generation and recognition for highly complex and interdependent concepts. Admittedly, \backronym is a small scale dataset with green screen background, which restricts the models trained on it to have specific biases. Nonetheless, this is the first attempt to understand how fine-grained states (cut styles) can be transferred to diverse objects. We explore this by using \backronym as a test set for larger models, fine-tuning these models using \backronym and trying them with or without a green screen background. We further see the potential of using \backronym for benefiting the community in even more challenging tasks such as 3D reconstruction, video frame interpolation, state change generation, \etc.

\section{Conclusion} In this paper, we propose \backronym, a new dataset for measuring the ability of models to recognize and generate unseen compositions of objects in different states, a skill known as compositional generalization. We also introduce two tasks, Compositional Image Generation and Compositional Action Recognition, and benchmark the performance of state-of-the-art generative models and video recognition methods on these tasks. We show the challenges with the existing approaches and their failure in some cases in their ability to generalize to new compositions. However, these two tasks are just the tip of the iceberg. Understanding object states is important for multiple image and video tasks such as 3D reconstruction, future frame prediction, video generation, summarization, and parsing of long-term video. We hope that our dataset will help the computer vision community to propose and learn new compositional tasks for images, videos, 3D, and beyond.\looseness=-1

\smallskip
\noindent\textbf{Acknowledgements.} The authors would like to dedicate this paper to the memory of Vinoj Jayasundara. His creativity, contributions and enthusiasm for the field of Computer Vision will continue to inspire us. We would also like to thank Snehesh, Chahat, Kanishka, and Pulkit for their valuable conversations during data collection. This work was partially funded by DARPA SAIL-ON (W911NF2020009) program and NSF CAREER Award (\#2238769) to AS.

\newpage
{\small
\bibliographystyle{ieee_fullname}
\bibliography{egbib}
}
\newpage
\appendix
\section*{Appendix} 
\begingroup
\let\clearpage\relax
\tableofcontents
\endgroup
\section{Scope and Limitations}
The objective of ChopNLearn dataset is compositional generation and recognition, using a granular and structural understanding of transferable object states. Terms such as `slice', `dice' alone often lead to a loss of granular information. For \eg, a sliced apple can be horizontally or vertically sliced, or cut in the half, and then sliced as semi-circles. Hence, we use more specific categories than other traditional state change datasets as shown in Tab. 1. Moreover, the subtle state change understanding is a challenging task on its own merit~\cite{fathi, look_c,multi}. Recognizing/segmenting actions in a video is a complementary task and an interesting future direction, but is currently beyond the scope of this work.
\begin{figure}[t]
\centering
\includegraphics[width=0.89\linewidth]{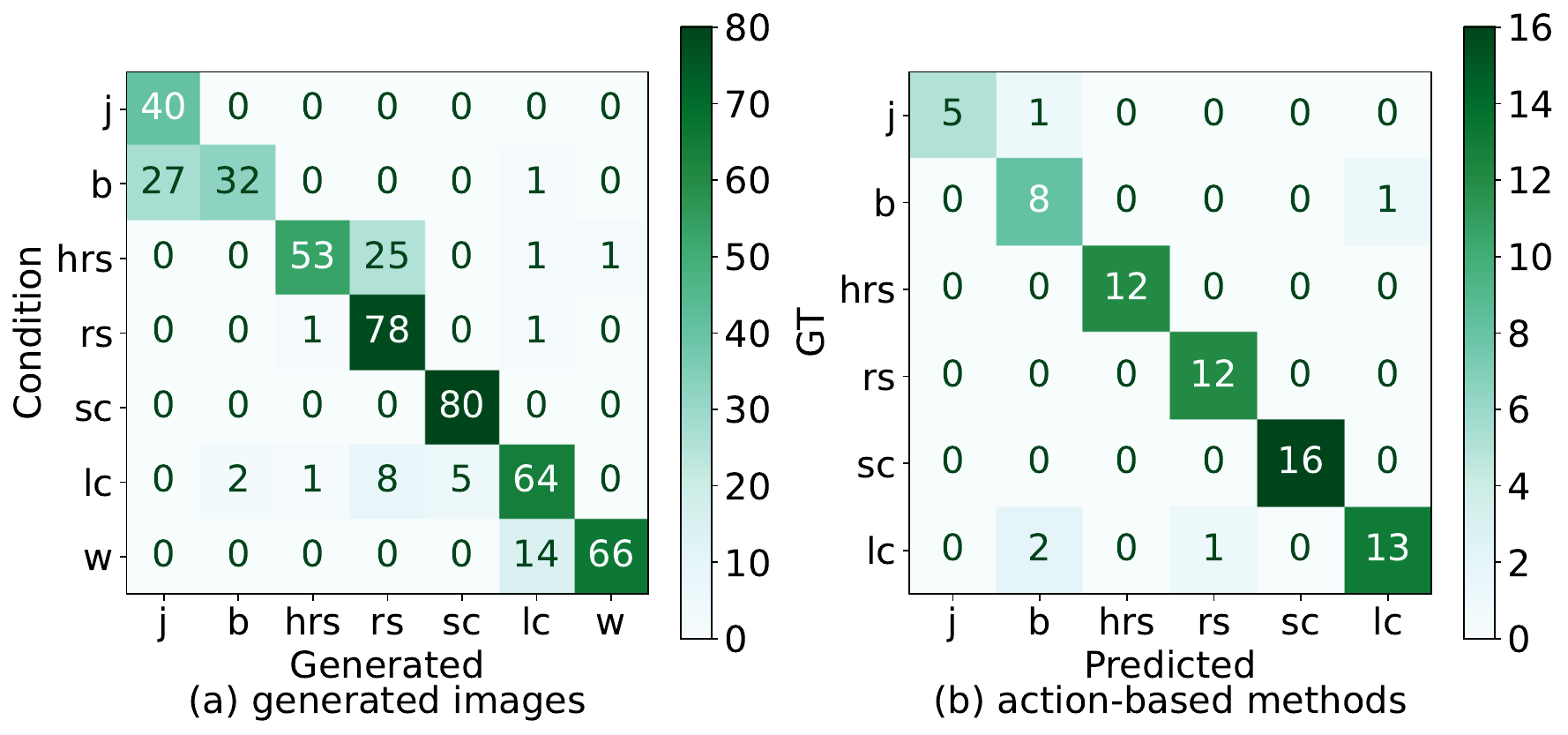}
\caption{Confusion matrix for generation \& action-based tasks.}
\label{fig:confusion_matrix}
\vspace{-1.2em}
\end{figure}

Moreover, we acknowledge that making the classes more granular can be confusing for the model, which appears similar. To confirm this, in Fig.~\ref{fig:confusion_matrix}, we show the confusion matrix for generated images (classified by the StateClassifier, and action-based method using the final states (CAF+R3D for Split 3 in {Tab. 3}). We see baton and julienne, half round slice and round slice, are two difficult pairs for compositional generation. In contrast, the action-based method can classify most states correctly. We hypothesize that since action-based methods use trajectory, and multiple frames for classification, the confusion between similar object-state pairs is significantly reduced.
\begin{figure*}[h!]
    \centering
    \includegraphics[width=\linewidth]{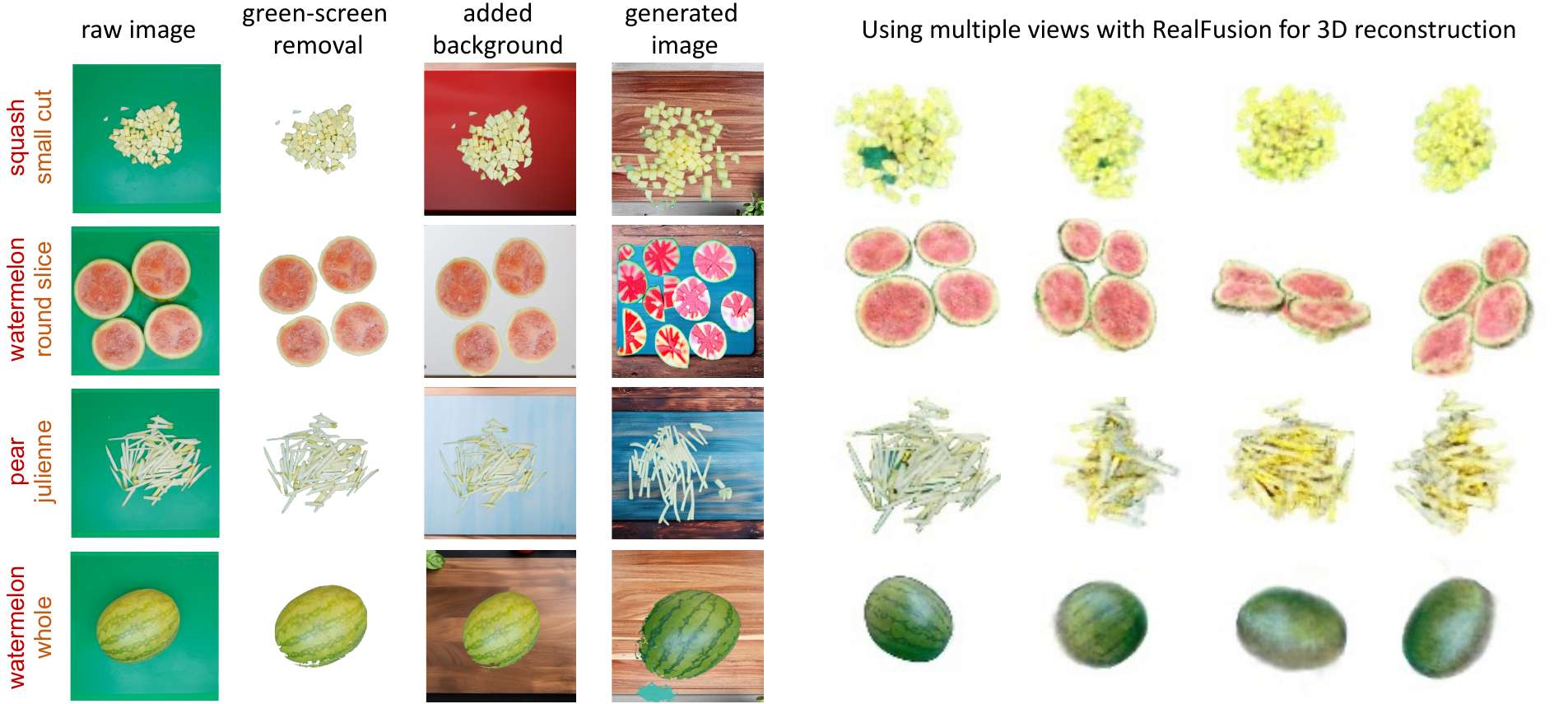}
    \caption{We show different uses of our dataset. The first column shows the raw images. The second column shows basic Python based green screen removal techniques on the dataset. The third column uses generated kitchen chopping board images replacing the green screen using the segmented object pieces. The fourth column shows results of training stable our benchmark SD+TI+FT model with the images without a green screen. Further the rightmost four columns show promising results towards 3D reconstruction of deformable objects, which can be a potential future research problem ChopNLearn can be used for.}
    \label{fig:exampels_3d}
\end{figure*}
\section{Future Work}
\label{sec:disc1}
\subsection{Green Screen Removal (extension).} As described in Section 4.3 of main paper, we chose green screen to focus on the object states, and such that the object pieces can be segmented easily. As some preliminary work, in~\Cref{fig:exampels_3d}, shows some results on how thresholding using simple opencv library functions works for segmenting the object pieces after styles of cuts are applied. Further, we use Midjourney~\cite{midjourney}, which is a Stable Diffusion~\cite{ho2022classifier} based text to image generator tool, to generate a set of images with chopping boards. For each camera angle, we generate 7-10 images, for red, yellow, blue, white and wooden chopping boards. The captions used for this explain the color of board, the view or angle and the surrounds, for instance ``empty plastic red colored chopping board in from this view point --style raw " is one of the captions used for one view.
Nonetheless, we are aware that achieving precise camera angles during image generation is a challenging task. Many of these models exhibit a bias toward placing certain fruits or vegetables around or atop the chopping board. As a result, we have occasionally supplemented the model with a reference image from the dataset, accompanied by a directive to "generate the same viewpoint and camera angle relative to the chopping board." Despite these efforts, the outcomes we achieve remain suboptimal, particularly when it comes to three-dimensional perspectives. The top view seems to yield the most favorable outcomes for background generation. We utilize these images to substitute the backgrounds in segmented images. However, there are instances where the generated background's view and angle do not align with those of the actual image from which the object segments originated. This occasionally results in sections of the objects appearing to levitate above the chopping board.
While we acknowledge that the authenticity of these generated images may sometimes fall short due to occasional inaccuracies in green screen removal and misalignment between object positioning and background image angles, this experiment still holds value in mitigating green screen bias during model training. Additionally, our analysis with respect to green screen removal and background addition sheds light on a significant prospective challenge – the ability to mat and position objects convincingly from diverse angles within backgrounds to achieve a realistic effect. 
Hopefully, paves the way for new avenues of research and potential applications of ChopNLearn in the realm of detailed background matting.

\subsection{3D reconstruction}
Collecting data as well as generating 3D models for deformable objects is still an open problem. We demonstrate results of some preliminary experiments with our dataset for this task. We use RealFusion \cite{melaskyriazi2023realfusion} to recover a promising 3D scene from a single image of our various cut states~\Cref{fig:exampels_3d}. We believe that with our multi-view camera setup, this direction is worth exploring in future work for more accurate 3D reconstruction and can be an interesting task.
\begin{figure}[h]
    \centering
    \includegraphics[width=.95\linewidth]{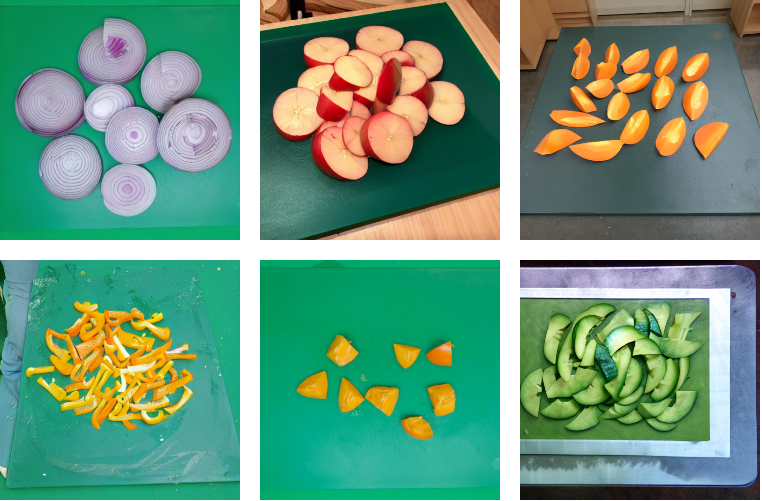}
    \caption{Examples of images from the test set and samples from the generative models presented to participants in the user study}
    \label{fig:userstudy}
\end{figure}
\begin{figure}[h]
    \centering
    \includegraphics[width=0.8\linewidth]{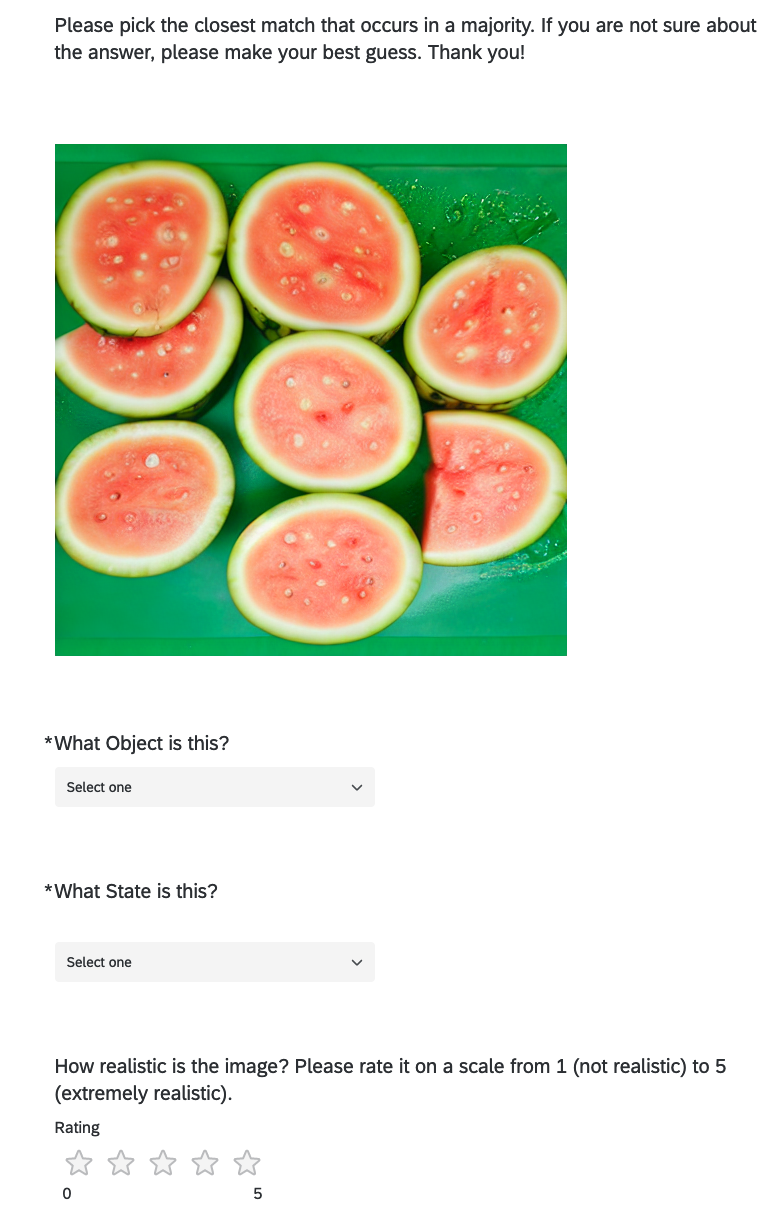}
    \caption{Snapshot of questionnaire presented to participants of the user study}
    \label{fig:userstudy3}
    \vspace{-1em}
\end{figure}

\section{Details of User Study}
The purpose of conducting a user study was to see if our generative models were able to create images that were of high fidelity and stayed true to capturing the semantic understanding of the object-state composition provided as a text prompt. We chose 20 compositions from the test set, which are unseen as a pair in the training and finetuning of the generative models. These compositions from the test set are also given as a text prompt to five generative models, i) Dreambooth ii) Stable Diffusion iii) Stable Diffusion+Textual Inversion iv) Stable Diffusion finetuned on our training dataset v) Stable Diffusion + Textual Inversion finetuned on our training dataset.  We evenly chose a distribution of 5 samples per composition, and including the test set + 5 generative models, we had 6 sets to sample images from. The total number of images used for the study was 750 and we asked 30 participants to label each of these images for their object and state as well as rate the realism on a scale of 1-5. We show some examples of images encountered in our user study in Figure~\ref{fig:userstudy} and a snapshot of how the user study questionnaire looks like in Figure ~\ref{fig:userstudy3}.
\section{Compositional Image Generation}
\subsection{Dataset Split}

\begin{figure}[htbp]
    \centering
    \includegraphics[width=.95\linewidth]{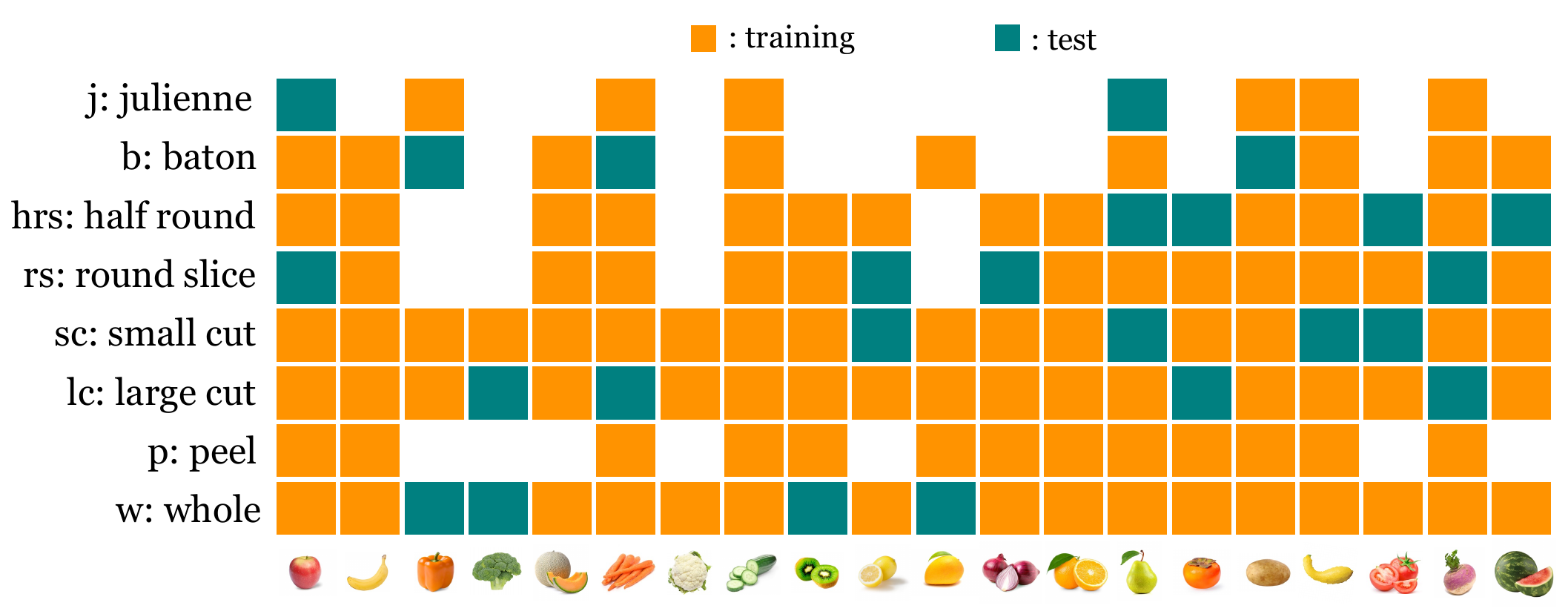}
    \caption{\textbf{Dataset Split used in The Compositional Image Generation Task}. Training compositions and test compositions are marked with orange and teal, respectively. Unmarked compositions are not included in our dataset. }
    \label{fig:gen_split}
\end{figure}

\begin{figure*}[h!]
    \centering
    \includegraphics[width=\linewidth]{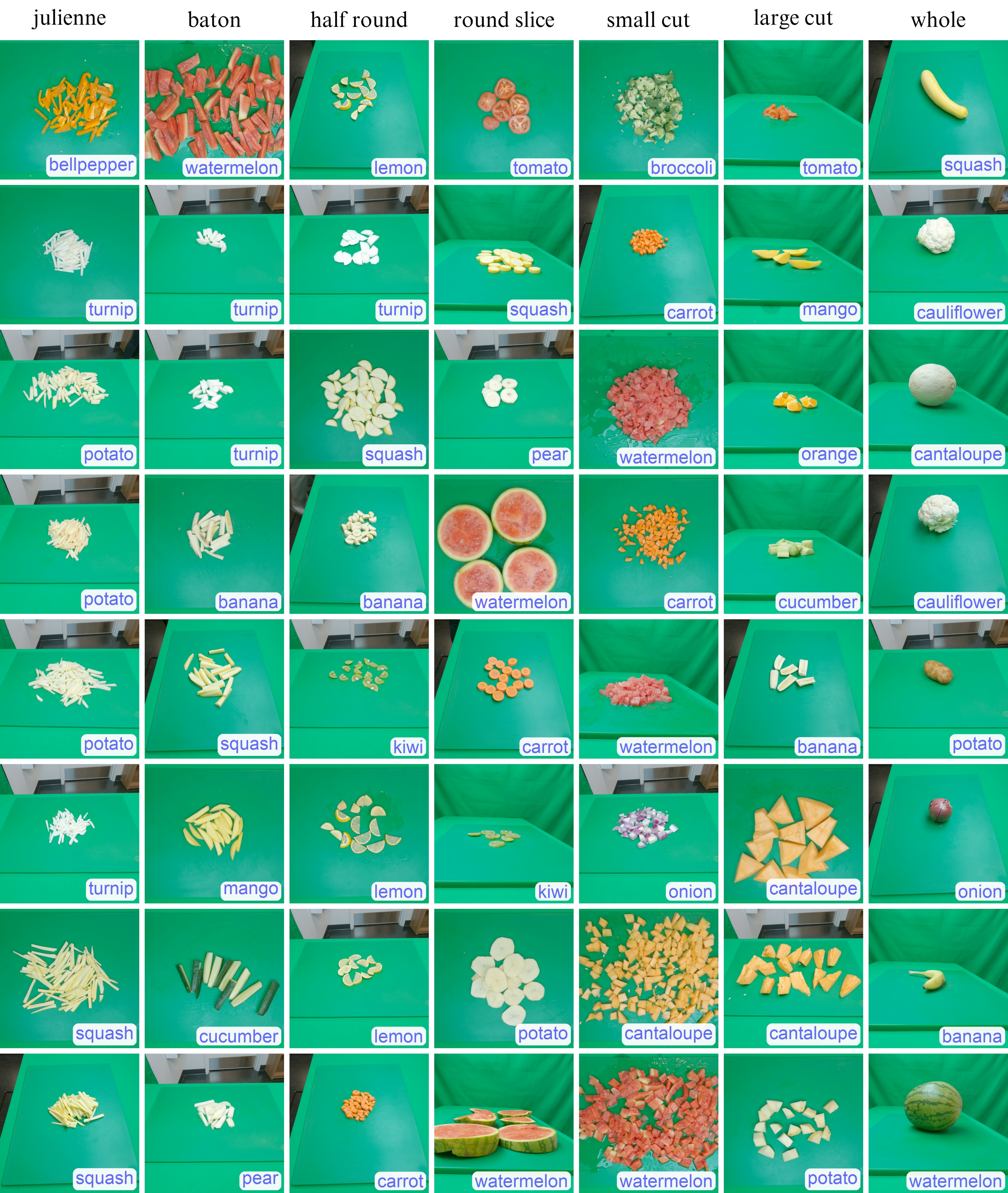}
    \caption{\textbf{Example Images In The Training Set}. Eight example images are shown in a column for each state. State labels are shown in the first row. Object labels are marked on the bottom right corner of each image.}
    \label{fig:exampels_train}
\end{figure*}

\begin{figure*}[h!]
    \centering
    \includegraphics[width=\linewidth]{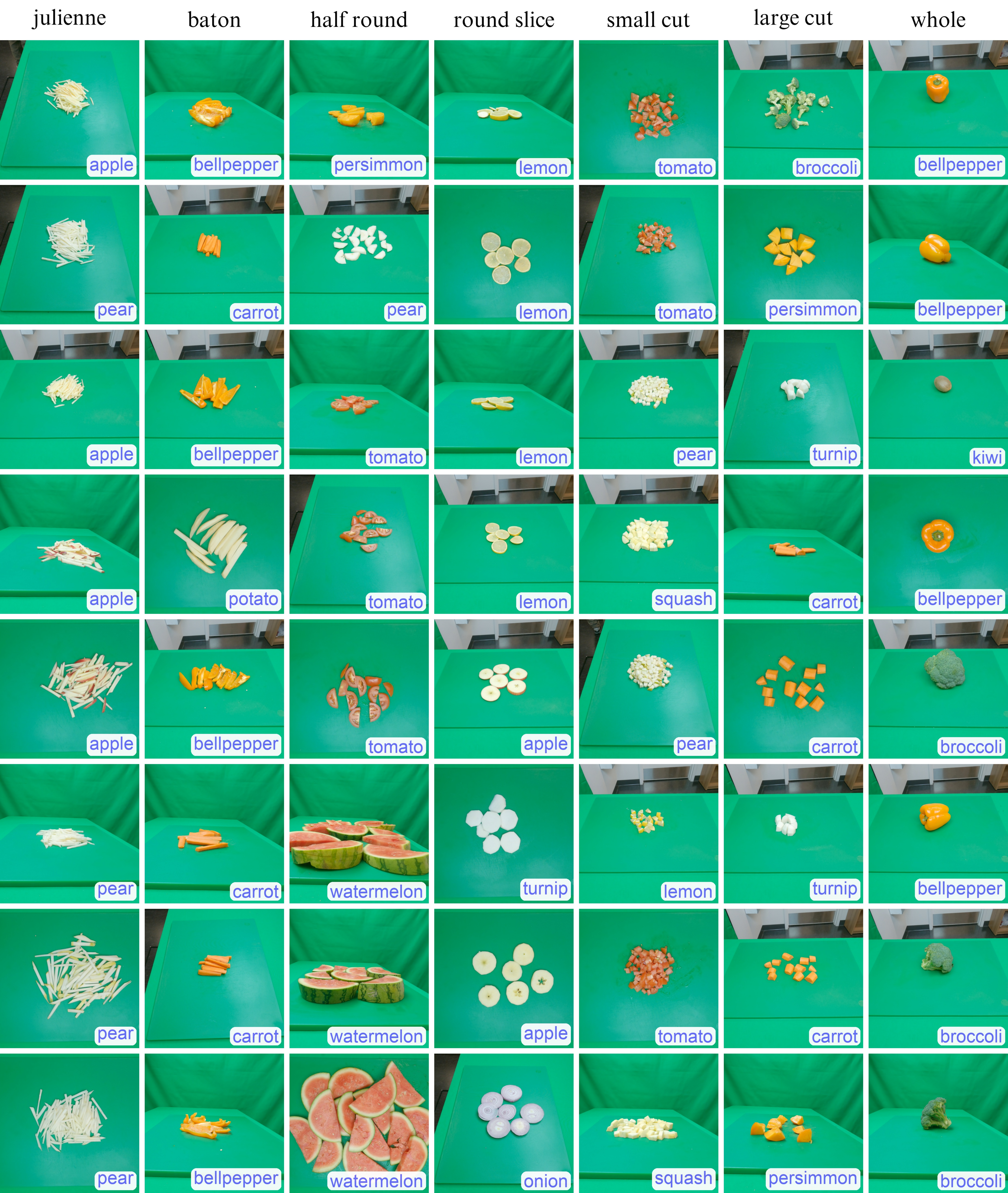}
    \caption{\textbf{Example Images In The Test Set}. Eight example images are shown in a column for each state. State labels are shown in the first row. Object labels are marked on the bottom right corner of each image.}
    \label{fig:exampels_test}
\end{figure*}

In the compositional image generation task, we split all (object, state) compositions into a training set consisting of 87 compositions and a test set consisting of 25 compositions. For each composition of object and state present in the test set, the training set includes exactly one of either the object, or the state, but not both. We also ensure that for each (object, state) composition $(o, s_i)$ in the test set, there exists a composition $(o, s_j)$ in the training set, where $s_i$ and $s_j$ belong to the same state related group defined in Section~3 of our main paper. Each combination in our dataset has 8-12 images, resulting in a total of 1032 images in the training set and 296 images in the test set. \cref{fig:gen_split} illustrates the detailed dataset split used in the compositional image generation task. In this figure, training compositions and test compositions are marked with orange and teal, respectively. Unmarked compositions are not included in our dataset. \cref{fig:exampels_train} and \cref{fig:exampels_test} show some example images in our training set and test set, respectively. 
\subsection{Number of views.}
We assess the impact of the number of views on the image generation task in Tab.~\ref{tab:view_ab} using the SD+FT+TI setting. Using more views improves training data in terms of both quantity and diversity, yielding results with better patch FID and object accuracy, and maintaining high state accuracy even though generating images in more views is more difficult. 
The use of 4 cameras also has applications in few-shot 3D reconstruction tasks, which although beyond the scope of current work, are discussed in~\Cref{sec:disc1}.
\begin{table}[htbp]
  \centering
  \vspace{-0.1in}
  \caption{\footnotesize \textbf{Number of views ablation results.}}
  \label{tab:baseline_results}
  \resizebox{0.4\textwidth}{!}{
    \begin{tabular}{@{}lccccc@{}}
      \toprule
      View IDs & Object Acc. (\%) $\uparrow$ & State Acc. (\%) $\uparrow$  & Patch FID $\downarrow$ \\
      \midrule
      1 & 42.4 & 78.2 & 184.7 \\
      1, 2 & 56.8 & 81.2 & 121.4 \\
      1, 2, 3 & 66.2 & 78.3 & 115.4 \\
      1, 2, 3, 4 & 67.8 & 81.4 & 82.2 \\
      \bottomrule
    \end{tabular}
  }
  \label{tab:view_ab}
  \vspace{-0.2in}
\end{table}
\subsection{Patch FID Details}
We propose patch FID to access the quality of the generated images. In short, it calculates Fr\'echet Inception Distance on the image patch level.
Specifically, we modify the standard FID by sampling $224 \times 224 $ random crops from the real images, as well as the synthetic images. We use 32 patches per image. For each generative model, we compute patch FID using all available real image patches and 16000 generated image patches, and report the average number for the test compositions.

\begin{table*}[t]
\centering
\caption{Input frames ablation. We do experiments on two settings to demonstrate that taking full video as input is necessary. The first row takes the full length of the video as input. The second row takes the first and last frames of the video as input. Object-final state classification accuracy is reported here.} 
\label{tab:res_input_ab}
\footnotesize
\renewcommand{\tabcolsep}{3pt}
\resizebox{0.8\textwidth}{!}{
    \begin{tabular}{@{}lllcccccccc@{}}
    \toprule
    & & & \multicolumn{2}{c}{Split 1}
    & & \multicolumn{2}{c}{Split 2}
    & & \multicolumn{2}{c}{Split 3}\\
    \cmidrule{4-5}
    \cmidrule{7-8}
    \cmidrule{10-11}
    Input & Model &  Features & acc@1 & acc@3 & & acc@1 & acc@3 && acc@1 & acc@3  \\
    \midrule
    Full  & Transformer~\cite{vaswani2017attention} & I3D~\cite{i3d} & 10.9 & 44.3 && 14.6 & 44.2 && 11.1 & 44.4\\
    First\&Last & Transformer~\cite{vaswani2017attention} & I3D~\cite{i3d} & 6.3 & 25.6 && 9.8 & 31.0 && 7.9 & 34.9\\
    \bottomrule
    \end{tabular}
}
\end{table*}
\subsection{Object State Classifier Details}
As mentioned in our main paper, to automatically evaluate the correctness of 
the generated images, we train a classifier on real images for classifying objects and states independently. This classifier is built on a CLIP-ViT-B/32~\cite{clip}. To classify an input image, it takes this image and texts of all possible labels (all objects or all states) as input. Cosine similarities between the image embedding and text embeddings of all possible labels are computed as the classification logits, which are used to calculate the standard cross entropy loss for classification problems. During hyperparameter-searching, we fine-tune the CLIP model on a different training split that all (object, state) compositions are seen, and report the validation accuracies in the Table~2 of our main paper. One single model is used to predict both object and state. 

After deciding on all hyperparameters and training settings, we train our final-version object state classifier on all available data in our dataset to maximize its performance. We keep all parameters in the CLIP model learnable and train it $2000$ epochs using a learning rate of $3e-5$. We use a batch size of $128$, and a warm-up cosine learning rate schedule~\cite{loshchilov2016sgdr}.

\subsection{Method Details}
\noindent
\textbf{Stable Diffusion. (SD)}
We briefly describe classifier-free guidance in diffusion models. Diffusion models generate an image from Gaussian noise via an iterative denoising process. Expected mean square error is used as the denoising objective:
\begin{align}
    \mathcal{L}_\text{Diff} = \mathbb{E}_{\mathbf{x}_0, \boldsymbol{\epsilon}, t\sim \mathcal{U}(0, 1)} \left[ \parallel \boldsymbol{\epsilon} - \mathbf{\epsilon}_\theta(\alpha_t \mathbf{x}_0 + \sigma_t \boldsymbol{\epsilon}, \mathbf{c})  \parallel^2 \right]
\label{eq:diffusion}
\end{align}
\noindent
where $\mathbf{x}_0$ is an image and $\mathbf{c}$ is the optional condition from the training data. $\boldsymbol{\epsilon}$ is the additive Gaussian noise. $\alpha_t, \sigma_t$ are scalar functions of time step $t$.  $\boldsymbol{\epsilon}_\theta$ is the diffusion model with trainable parameters $\theta$. 
For sampling images from the text condition, SD employs classifier-free guidance~\cite{ho2022classifier}, such that at every time step (during sampling), predicted noise is adjusted via:
\begin{align}
    \hat{\boldsymbol{\epsilon}_\theta}(\mathbf{x}_t, \mathbf{c}) = \omega \boldsymbol{\epsilon}_\theta(\mathbf{x}_t, \mathbf{c}) + (1-\omega)\boldsymbol{\epsilon}_\theta(\mathbf{x}_t)
\label{eq:guidance}
\end{align}
\noindent
where $\omega$ is the guidance scale. In our experiments, $\omega$ is set to be $7.5$ in all methods using it.

\noindent
\textbf{SD + Textual Inversion (TI).} In this method, \Cref{eq:diffusion} is used for token embedding optimization.  SD weights are kept fixed during training.
We use a learning rate of $3e-3$ with a warm-up cosine learning rate schedule~\cite{loshchilov2016sgdr}, a batch size of 4, and train the model for 16000 steps.

\noindent
\textbf{DreamBooth.}
The text prompt we used for DreamBooth fine-tuning is ``An image of $o_i$ cut in the [V] style", where $o_i$ is the $i^{th}$ object and [V] is a rare unique identifier representing the state this model is fine-tuned for.
The goal of DreamBooth is to overfit a small dataset without drifting too far away from the pre-trained model.
Following the available open-source implementation,
we use a fixed learning rate of $5e-6$, a batch size of 1, and train the model for 400 steps.
\begin{table*}[t]
\centering
\caption{Other splits: We also present other possible splits of data. All the results are using I3D~\cite{i3d} pre-trained features along with one layer Transformer~\cite{vaswani2017attention} model. Comp. represents the initial: object-initial state composition and final: object-final state composition results for each split.} 
\label{tab:2}
\footnotesize
\renewcommand{\tabcolsep}{3pt}
\resizebox{0.95\textwidth}{!}{
    \begin{tabular}{@{}lccccccccccccccccc@{}}
    \toprule
     & \multicolumn{2}{c}{Split 4}
    & & \multicolumn{2}{c}{Split 5}
    & & \multicolumn{2}{c}{Split 6}
    & & \multicolumn{2}{c}{Split 7}
    & & \multicolumn{2}{c}{Split 8}
    & & \multicolumn{2}{c}{Split 9}
    \\
    \cmidrule{2-3}
    \cmidrule{5-6}
    \cmidrule{8-9}
    \cmidrule{11-12}
    \cmidrule{14-15}
    \cmidrule{17-18}
     Comp. & acc@1 & acc@3 & & acc@1 & acc@3 && acc@1 & acc@3 && acc@1 & acc@3 && acc@1 & acc@3 && acc@1 & acc@3  \\
    \midrule
    Initial & 46.5 & 78.9 && 65.2 & 90.6 && 37.5 & 70.8 && 41.1 & 76.0 && 47.2 & 71.5 && 41.2 & 78.3\\
    Final & 45.3 & 75.9 && 73.9 & 92.6 && 37.7 & 69.2 && 41.4 & 72.2 && 48.9 & 73.0 && 42.9 & 77.6 \\
    \bottomrule
    \end{tabular}
}
\end{table*}
\begin{table*}
\vspace{-0.05in}
\centering
\caption{Results of finetuning R3D~\cite{r3d} backbone. ``Start/End" denotes the prediction results for the initial and the final state composition with the correct object type.} 
\label{tab:res_finetune}
\footnotesize
\renewcommand{\tabcolsep}{3pt}
\resizebox{0.95\textwidth}{!}{
    \begin{tabular}{@{}lccccccccccccccc@{}}
    \toprule
    & & \multicolumn{4}{c}{Split 1}
    & & \multicolumn{4}{c}{Split 2}
    & & \multicolumn{4}{c}{Split 3}\\
    \cmidrule{3-6}
    \cmidrule{8-11}
    \cmidrule{13-16}
    & & \multicolumn{2}{c}{Start} & \multicolumn{2}{c}{End} 
    & & \multicolumn{2}{c}{Start} & \multicolumn{2}{c}{End}
    & & \multicolumn{2}{c}{Start} & \multicolumn{2}{c}{End}\\
    \cmidrule{3-4}
    \cmidrule{5-6}
    \cmidrule{8-9}
    \cmidrule{10-11}
    \cmidrule{13-14}
    \cmidrule{15-16}
    Model &  Finetune & acc@1  & acc@3 & acc@1  & acc@3 & & acc@1  & acc@3 & acc@1  & acc@3 && acc@1  & acc@3 & acc@1  & acc@3  \\
    \midrule
    CAF~\cite{stlt} &  & 53.5 & 88.7 & 57.8 & 88.7 & & 55.1 & 95.7 & 58.0 & 95.7 & & 62.0 & 93.0 & 63.4 & 93.0\\
    \midrule
    CAF~\cite{stlt} & \checkmark & 80.3 & 98.6 & 87.3 & 98.6 & & 84.1 & 98.5 & 89.9 & 98.5 & & 88.7 & 98.6 & 88.7 & 98.6\\
    \bottomrule
    \end{tabular}
}
\vspace{-0.1in}
\end{table*}

\noindent
\textbf{SD + Fine-tuning (FT).}
We also fine-tune SD while keeping the text encoder fixed. The UNet parameters of the diffusion model are optimized using the diffusion loss defined by ~\Cref{eq:diffusion}.
We use a learning rate of $5e-6$ with a warmup cosine learning rate schedule~\cite{loshchilov2016sgdr}, a batch size of 4, and train the model for 8000 steps.

\noindent
\textbf{SD + TI + FT.} When combining SD fine-tuning and Textual Inversion~\cite{gal2022image} together, we use a learning rate of $5e-6$ for all UNet parameters and a learning rate of $3e-3$ for all added token embeddings. A warmup cosine learning rate schedule~\cite{loshchilov2016sgdr} is employed for all parameters. We use a batch size of 4, and train the model for 16000 steps.

\subsection{Additional Qualitative Results}
\begin{figure*}[h]
\centering
    \includegraphics[width=\linewidth]{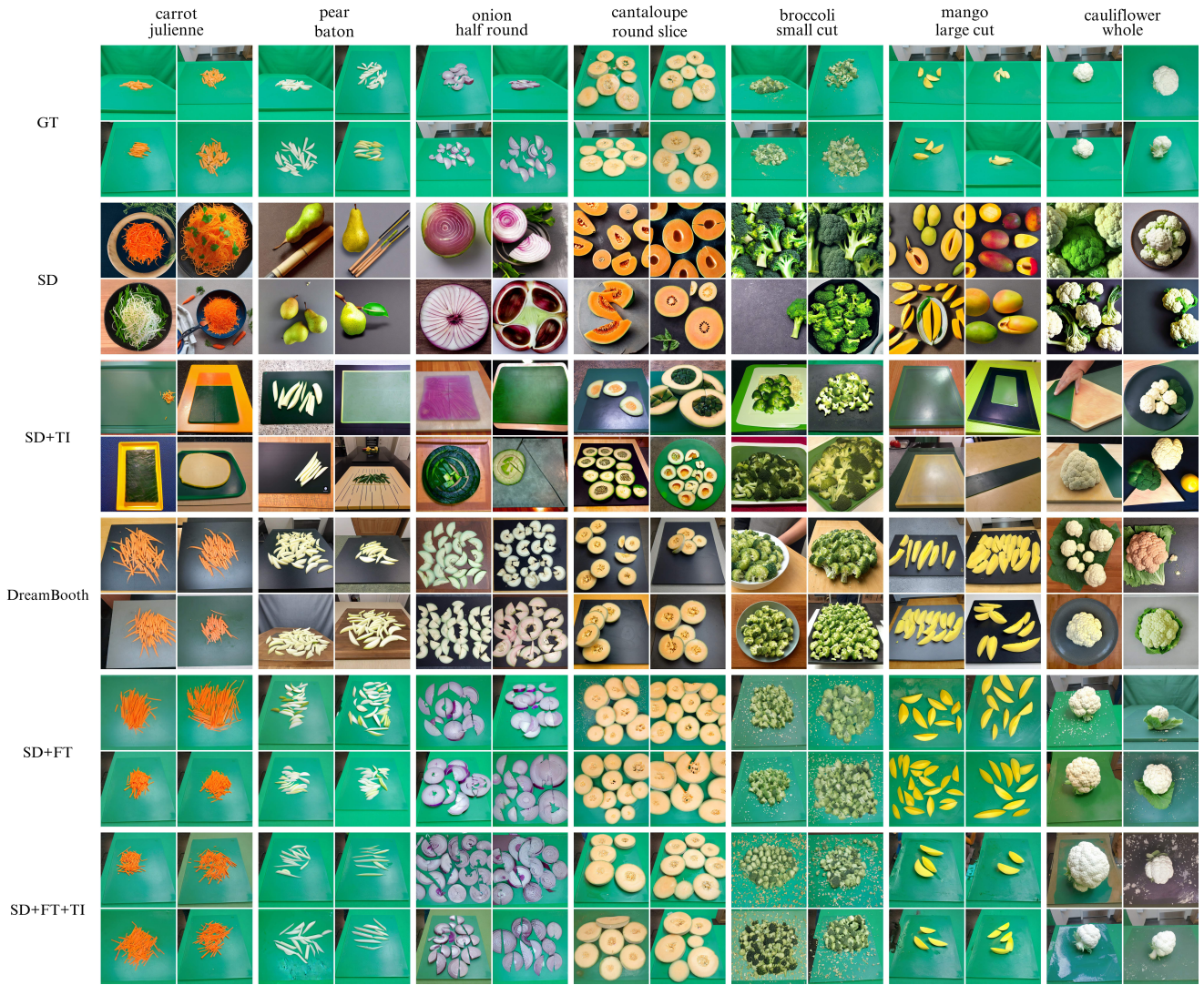}
    \caption{\textbf{Additional Compositional Generation Samples Using Training Compositions} Ground Truth (GT) real images are shown in the first row for reference. Seven object-state compositions in the training set are displayed, each with four generated samples for each method. Please zoom in to see the details.} 
\label{fig:comp_supp_train}
\end{figure*}

\begin{figure*}[h]
\centering
    \includegraphics[width=\linewidth]{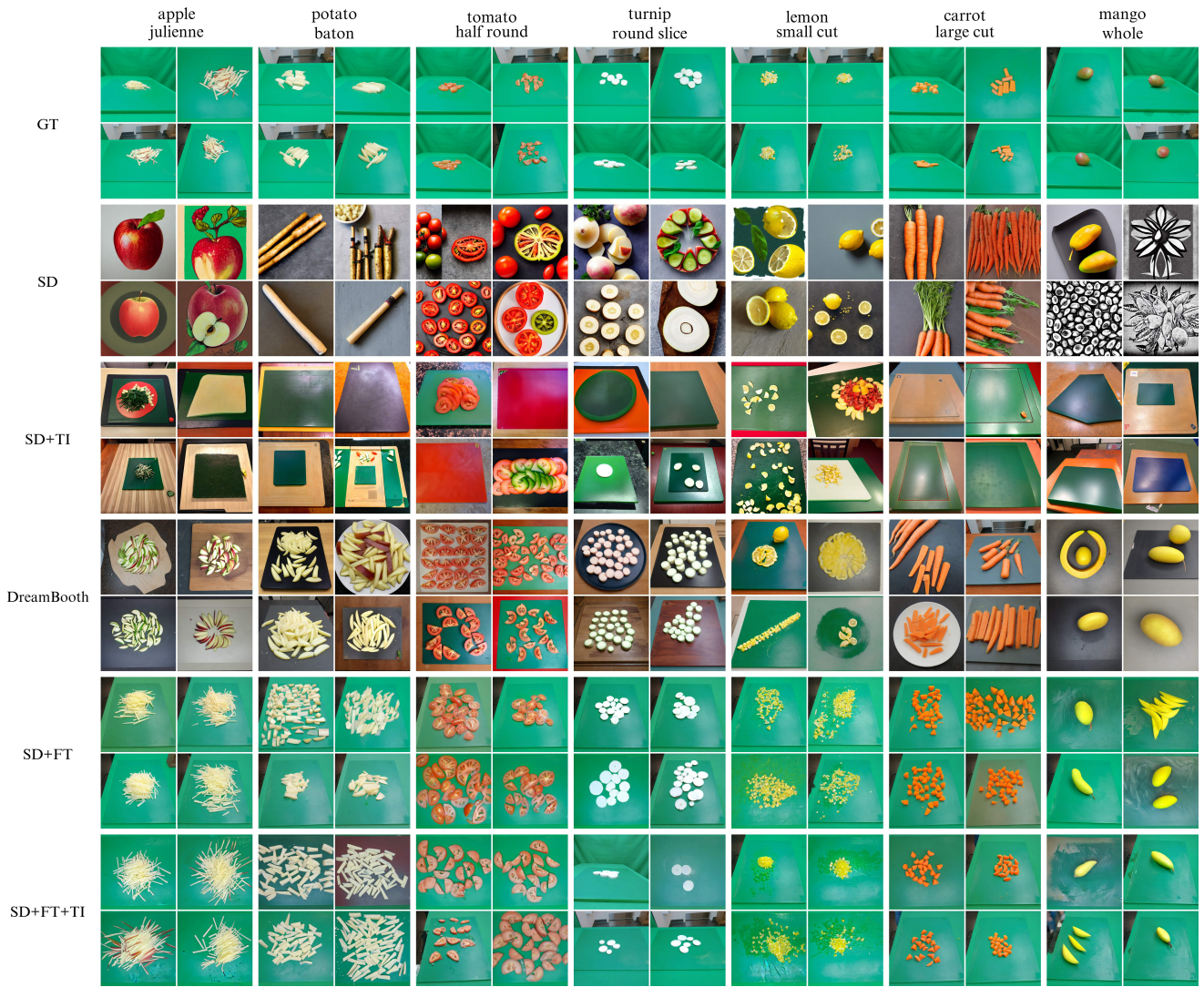}
    \caption{\textbf{Additional Compositional Generation Samples Using Test Compositions} Ground Truth (GT) real images are shown in the first row for reference. Seven object-state compositions in the test set are displayed, each with four generated samples for each method. Please zoom in to see the details.} 
\label{fig:comp_supp_test}
\end{figure*}

To better compare the compositional image generation performance of various methods discussed in the main paper, we show additional generated images from them for seven (object, state) compositions in \cref{fig:comp_supp_train} and \cref{fig:comp_supp_test}, where the compositions are from the training set and test set, respectively. 

\section{Compositional Action Recognition} 
\subsection{Dataset Splits}
Given the diversity of views and object types and styles, we can construct multiple training and testing splits. In this paper, we present results on three selected splits. For each split, we create training, test and validation set. The validation set is for evaluating the model on training classes, which consists of 10-15\% unseen samples for the seen training compositions. Training and test sets have a disjoint set of compositional classes, in an 80-20\% ratio. All of the splits in our dataset are created based on object-final state compositions in the videos. 

We leverage these related groups defined in Section 3.1 in the main paper, to create different splits for training and testing. All splits use multi-view camera angles and involve creating seen and unseen object-final state compositions in training and testing sets. This ensures cross-view training and test splits, as used in other multi-view datasets~\cite{ntu-rgb, multi-view_act}. The training set consists of samples from three cameras, while the test set includes samples of compositions from one camera whose view is never seen during training.

Similar to the three splits mentioned in the paper, we explore multiple other splits, with different constraints to choose those 3 splits. We find that other splits were not as challenging for the existing baselines, and hence only propose 3 splits that are challenging. In~\Cref{tab:2}, we present the results for splits 4-9, which were considered for the data. We only show I3D~\cite{i3d} based Transformer model for these splits. All the splits consider the constraints for the object-final state. The details of each split are as follows:

\noindent \textbf{Split 4:} This is the same as the split used for the Compositional Image Generation task (mentioned in Section 4 of the main paper). We use the related groups to split the object-final states, such that objects which are seen with one of the states in a related group in training, are tested on the other related group during testing. This is also similar to Split 2 in the main paper, however, the multi-view constraint is not there. All the camera views are used for training and testing.

\noindent \textbf{Split 5:} In this split, we have the participant constraint. All samples from participants 1 and 2 are part of the training set, while samples from participant 3 are in the test set.

\noindent \textbf{Split 6:} This is a combination of split 4 and 5, which has two constraints: using related groups for splitting object-final states in different splits, and using only participant id 3 for the test set.

\noindent \textbf{Split 7:} This split is about multi-camera view. We use Camera 1,2,3 views in the training set, while the camera 4 view is part of the test set. No other constraint regarding related groups for splitting on the basis of object-final states is used.

\noindent \textbf{Split 8:} This split is similar to Split 1 in the main paper, without the multi-camera constraint. The object-final state compositions are split randomly into train/test. We use all camera views for both sets, without constraining to distinct views for each set.

\noindent \textbf{Split 9:} This split is similar to Split 3 in the main paper, without the multi-camera constraint. The object-final state compositions are split based on random groups for objects and states. We use all camera views for both sets, without constraining to distinct views for each set.\\
We do not have a split having all constraints, \ie participant constraint, related groups and multi-view constraint, since all of these together end up leaving a total of 400 video clips, which are very few for training and testing. We show only top@1 accuracy for object-final state composition in \Cref{tab:2}.

\subsection{First and last segment classification}
For compositional action recognition, we emphasize that the model must learn to predict the object-initial state composition and the object-final state composition. 
Moreover, some works~\cite{fathi,nirat} use a similar setup for object state classification and use only the first and last frame/segment for this. Ideally, the first few frames and last few frames should be sufficient for understanding the changes in object states. We also experiment with the first and last segments of videos, for classification. The results for the 3 selected splits (mentioned in the main paper) are in~\Cref{tab:res_input_ab}. We find that using the additional middle frames improves the classification accuracy for the final composition. 

\subsection{Finetuning Backbone} The results we show in the paper without finetuning any pre-trained features (I3D~\cite{i3d}, MIL-NCE~\cite{mil}, R3D~\cite{r3d}). For the sake of completeness, we also show results with finetuning the backbone for R3D features in~\Cref{tab:res_finetune}. 
Although the top@1 accuracy is much better, it is still not 100\%. Moreover, the dataset is much smaller and overfits very quickly for backbones which are trained on 10x more data. Hence, for sake of benchmarking, we propose not fine-tuning the features for consistency. 

\section{Project Webpage and License}
\noindent For more details, results and analysis, please visit our website at: 
\textcolor{magenta}{https://chopnlearn.github.io}.

\noindent \textbf{License.} All files in this dataset are copyright by us and published under the Creative Commons Attribution-NonCommerial 4.0 International License, found  at 
\textcolor{magenta}{https://creativecommons.org/licenses/by-nc/4.0/}. 
This means that you must give appropriate credit, provide a link to the license, and indicate if changes were made. You may do so in any reasonable manner, but not in any way that suggests the licensor endorses you or your use. You may not use the material for commercial purposes. \looseness=-1
\end{document}